\documentclass[twoside]{article}
\usepackage{ecj,palatino,epsfig,latexsym,natbib}
\usepackage{url}                                                          
\usepackage{lipsum}
\usepackage{graphicx} 
\usepackage{xcolor}
\usepackage{algorithm}
\usepackage[noend]{algpseudocode}
\usepackage{flushend}
\usepackage{fixltx2e}
\newcommand{\algoname}{TBR-Evolution}

\newif\iftodos

\todostrue

\iftodos
\newcommand{\todo}[1]{\textcolor{red}{[#1]}} 
\newcommand{\done}[1]{\textcolor{green}{[#1]}} 
\newcommand{\comment}[1]{\textcolor{blue}{[#1]}} 
\else
\newcommand{\todo}[1]{} 
\newcommand{\done}[1]{} 
\newcommand{\comment}[1]{} 
\fi

\begin{document}

\ecjHeader{x}{x}{xxx-xxx}{200X}{Evolving a Behavioral Repertoire for a Walking Robot}{A. Cully and J.-B. mouret}
\title{\bf Evolving a Behavioral Repertoire\\ for a Walking Robot}

\author{\name{\bf A. Cully} \hfill \addr{cully@isir.upmc.fr}\\
        \addr{Sorbonne Universit\'es, UPMC Univ Paris 06, UMR 7222, ISIR, F-75005, Paris, France}\\
        \addr{CNRS, UMR 7222, ISIR, F-75005, Paris, France}
\AND
       \name{\bf J.-B. Mouret} \hfill \addr{mouret@isir.upmc.fr}\\
        \addr{Sorbonne Universit\'es, UPMC Univ Paris 06, UMR 7222, ISIR, F-75005, Paris, France}\\
        \addr{CNRS, UMR 7222, ISIR, F-75005, Paris, France}
}

\maketitle

\begin{abstract}
Numerous algorithms have been proposed to allow legged robots to learn
to walk. However, the vast majority of these algorithms is devised to learn to walk
in a straight line, which is not sufficient to accomplish any
real-world mission.
Here we introduce the Transferability-based Behavioral Repertoire Evolution algorithm (\algoname{}), a novel evolutionary algorithm that
simultaneously discovers several hundreds of simple walking
controllers, one for each possible direction. By taking advantage of
solutions that are usually discarded by evolutionary processes, \algoname{} is substantially
faster than independently evolving each controller. Our technique
relies on two methods: (1) novelty search with local competition,
which searches for both high-performing and diverse solutions, and (2) the
transferability approach, which combines simulations and real tests to
evolve controllers for a physical robot. We evaluate this new technique on a hexapod
robot. Results show that with only a few dozen short experiments
performed on the robot, the algorithm learns a repertoire of
controllers that allows the robot to reach every point in its reachable
space. Overall, \algoname{} opens a new kind of learning algorithm
that simultaneously optimizes all the achievable behaviors of a robot.

\end{abstract}

\section{Introduction}
Evolving gaits for legged robots has been an important topic in evolutionary computation for the last 25 years \citep{deGaris1990,lewis1992genetic,kodjabachian1998evolution,hornby2005autonomous,Clune2011,Yosinski2011,sam14a}. That legged robots is a classic of evolutionary robotics is not surprising \citep{Bongard2013}: legged locomotion is a difficult challenge in robotics that evolution by natural selection solved in nature; evolution-inspired algorithm may do the same for artificial systems. As argued in many papers, evolutionary computation could bring many benefits to legged robotics, from making it easier to design walking controllers (e.g., \cite{hornby2005autonomous}), to autonomous damage recovery (e.g., \cite{bongard2006resilient,koos2013fast}). In addition, in an embodied cognition perspective \citep{Wilson2002,pfeifer2007body,Pfeifer2007}, locomotion is one of the most fundamental skills of animals, and therefore it is one of the first skill needed for embodied agents.

It could seem more confusing that evolutionary computation has failed to be central in legged robotics, in spite of the efforts of the evolutionary robotics community \citep{raibert1986legged,Siciliano2008}. In our opinion, this failure stems from at least two reasons: (1) most evolved controllers are almost useless in robotics because they are limited to walking in a straight line at constant speed (e.g. \cite{hornby2005autonomous,bongard2006resilient,koos2013fast}), whereas a robot that only walks in a straight line is obviously unable to accomplish any mission; (2) evolutionary algorithms typically require evaluating the fitness function thousands of times, which is very hard to achieve with a physical robot. The present article introduces \algoname{} (Transferability-based Behavioral Repertoire Evolution), a new algorithm that addresses these two issues at once.

Evolving controllers to make a robot walk in any direction can be seen as a generalization of the evolution of controllers for forward walking. A straightforward idea is to add an additional input to the controller that describes the target direction, then evolve controllers that use this input to steer the robot \citep{Mouret2006}. Unfortunately, this approach requires testing each controller for several directions in the fitness function, which substantially increases the time required to find a controller. In addition, an integrated controller that can use a direction input is likely to be more difficult to find than a controller that can only do forward walking. 

An alternate idea is to see walking in every direction as a problem of learning how to do many different -- but related -- tasks. In this case, an evolutionary algorithm could search for a \emph{repertoire of simple controllers} that would contain a different controller for each possible direction. This method circumvents the challenge of learning a complex controller and can be combined with high level algorithms (e.g. planning algorithms) that successively select controllers to drive the robot. Nevertheless, evolving a controller repertoire typically involves as many evolutionary processes as there are target points in the repertoire. Evolution is consequently slowed down by a factor equal to the number of targets. With existing evolution methods, repertoires of controllers are in effect limited to a few targets, because 20 minutes \citep{koos2013fast} to dozens of hours \citep{hornby2005autonomous} are needed to learn how to reach a single target.


\begin{figure}
\includegraphics[width=\columnwidth]{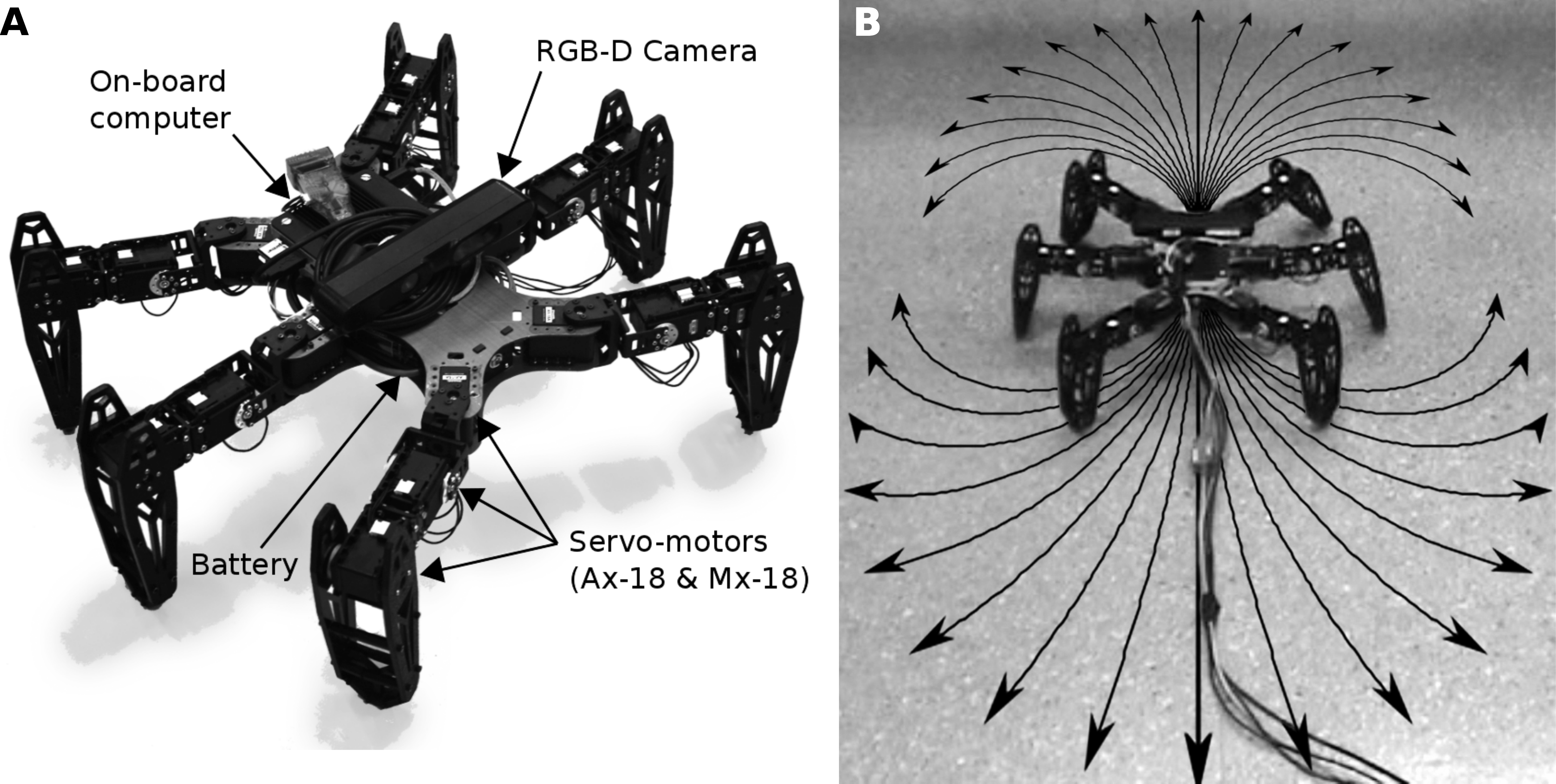}
\caption{(Right) The hexapod robot. It has 18 degrees of freedom (DOF), 3 for each leg. 
Each DOF is actuated by position-controlled servos (Dynamixel actuators).  
A RGB-D camera (Asus Xtion) is screwed on the top of the robot. The camera is used to estimate the forward displacement 
of the robot thanks to a RGB-D Simultaneous Localization And Mapping (SLAM) algorithm~\citep{endres12icra} from the ROS framework~\citep{Ros2009}.
(Left) Goal of TBR-Learning. Our algorithm allows the hexapod robot to learn to walk in every direction with a single run of the evolutionary algorithm.} 
\label{fig:objective}
\end{figure}


Our algorithm aims to find such a repertoire of simple controllers, but \emph{in a single run}. It is based on a simple observation: with a classic evolutionary algorithm, when a robot learns to reach a specific target, the learning process explores many different potential solutions, with many different outcomes. Most of these potential solutions are discarded because they are deemed  poorly-performing. Nevertheless, while being useless for the considered objective, these inefficient behaviors can be useful for other objectives. For example, a robot learning to walk in a straight line usually encounters many turning gaits during the search process, before converging towards straight line locomotion.

To exploit this idea, \algoname{} takes inspiration from the ``Novelty Search''algorithm \citep{lehman2011abandoning}, and in particular its variant the ``Novelty Search with Local Competition'' \citep{lehman2011evolving}. Instead of rewarding candidate solutions that are the closest to the objective, this recently introduced algorithm explicitly searches for behaviors that are different from those previously seen. The local competition variant adds the notion of a quality criterion which is optimized within each individual's niche. As shown in the rest of the present article, searching for many different behaviors during a single execution of the algorithm allows the evolutionary process to efficiently create a repertoire of high-performing walking gaits.


To further reduce the time required to obtain a behavioral repertoire for the robot, \algoname{} relies on the \emph{transferability approach} \citep{koos2011transferability,mouret2012}, which combines simulations and tests on the physical robot to find solutions that perform similarly in simulation and in reality. The advantages of the transferability approach is that evolution occurs in simulation but the evolutionary process is driven towards solutions that are likely to work on the physical robot. In recent experiments, this approach led to the successful evolution of walking controllers for quadruped \citep{koos2011transferability}, hexapod \citep{koos2013fast}, and biped robots \citep{oliveiraoptimization}, with no more than 25 tests on the physical robot.

We evaluate our algorithm on two sets of experiments. The first set aims to show that learning simultaneously all the behaviors of a repertoire is faster than learning each of them separately\footnote{This experiment is partly based on the preliminary results published in a conference paper \citep{cully2013behavioral}}. We chose to perform these experiments in simulation to gather extensive statistics. The second set of experiments evaluates our method on a physical hexapod robot (Fig.~\ref{fig:objective}, left) that has to walk forward, backward, and turn in both directions, all at different speeds (Fig.~\ref{fig:objective}, right). We compare our results to learning independently each controller. All our experiments utilize embedded measurements to evaluate the fitness, an aspect of autonomy only considered in a handful of gait discovery experiments~\citep{kimura2001reinforcement,hornby2005autonomous}.



%
%
%
%
%

\section{Background}

\subsection{Evolving Walking Controllers}
We call \emph{Walking Controller} the software module that
rhythmically drives the motors of the legged robot. We distinguish two
categories of controllers: \emph{un-driven controllers} and
\emph{inputs-driven controllers}. An un-driven controller always
executes the same gait, while an inputs-driven controller can change
the robot's movements according to an input (e.g. a speed or a
direction reference). Inputs-driven controllers are typically combined
with decision or planning
algorithms~\citep{russell2010artificial,currie1991plan,dean1991planning,kuffner2000rrt}
to steer the robot. These two categories contain, without
distinctions, both open-loop and closed-loop controllers and can be
designed using various controller and genotype structures.  For
example, walking gait evolution or learning has been achieved on
legged robots using parametrized periodic
functions~\citep{koos2013fast,chernova2004evolutionary,
  hornby2005autonomous, taraporecomparing,tarapore2014evolvability},
artificial neural networks with both direct or generative
encoding~\citep{Clune2011, valsalam2008modular,
  taraporecomparing,tarapore2014evolvability}, Central Pattern
Generators~\citep{kohl2004policy, ijspeert2007swimming}, or
graph-based genetic programming~\citep{filliat1999incremental,
  gruau1994automatic}.

When dealing with \emph{physical} legged robots, the majority of
studies only considers the evolution of un-driven walking controllers
and, most of the time, the task consists in finding a controller that
maximizes the forward walking speed
\citep{zykov2004evolving,chernova2004evolutionary,
  hornby2005autonomous,berenson2005hardware,
  Yosinski2011,mahdavi2006innately}. Papers on alternatives to
evolutionary algorithms, like policy gradients
\citep{kohl2004policy,tedrake2005learning} or Bayesian optimization
\citep{calandra2014experimental,lizotte2007automatic}, are also
focused on robot locomotion along a straight line.

Comparatively few articles deal with controllers able to turn or to
change the walking speed according to an input, especially with a
physical robot. Inputs-driven controllers usually need to be tested on
each possible input during the learning process or to be learned with
an incremental process, which significantly increases both the
learning time and the difficulty compared to learning an un-driven
controller. \cite{filliat1999incremental} proposed such a method, that
evolves a neural network to control a hexapod robot. Their neural
network is learned with several steps: first, the network is learned
in order to walk in a straight line; in a second step, a second neural
network is evolved on top of the walking controller be able to execute
turning manoeuvres. In a related task (flapping wing flight),
\cite{Mouret2006} proposed another approach, where an evolutionary
algorithm is used to design a neural network that pilots a simulated
flapping robot; the network was evaluated by its ability to drive the
robot to 8 different targets and the reward function was the sum of
the distances to the targets.

Overall, many methods exist to evolve un-driven controllers, while
methods for learning inputs-driven controllers are very time-expensive, difficult to apply on a physical robot, and require an
extensive amount of expert knowledge. To our knowledge, no current
technique is able to make a physical robot learning to walk in
multiple directions in less than a dozen hours.
In this paper, we sidestep many of the challenges raised by input-driven controllers while being able to drive a robot in every direction: we propose to abandon input-driven controllers, and, instead, search for a large number of simple, un-driven controllers, one for each possible direction.







\subsection{Transferability approach}

Most of the previously described methods are based on stochastic optimization algorithms that need to test a high number of candidate solutions. Typically, several thousands of tests are performed with policy gradient methods~\citep{kohl2004policy} and hundreds of thousands with evolutionary algorithms~\citep{Clune2011}. This high number of tests is a major problem when they are performed on a physical robot. An alternative is to perform the learning process in simulation and then apply the result to the robot. Nevertheless, solutions obtained in simulation often do not work well on the real device, because simulation and reality never match perfectly. This phenomenon is called \emph{the Reality Gap} \citep{jakobi1995noise,koos2011transferability}.
 
\emph{The transferability approach}~\citep{koos2011transferability,koos2013fast,mouret2012} crosses this gap by finding behaviors that act similarly in simulation and in reality. During the evolutionary process, a few candidate controllers are transferred to the physical robot to measure the behavioral differences between the simulation and the reality; these differences represent the \emph{transferability value} of the solutions. With these few transfers, a \emph{regression model} is built up to map solution descriptors to an estimated transferability value. The regression model is then used to predict the transferability value of untested solutions. The transferability approach uses a multi-objective optimization algorithm to find solutions that maximize both task-efficiency (e.g. forward speed, stability) and the estimated transferability value.

This mechanism drives the optimization algorithm towards solutions that are both efficient in simulation and transferable 
(i.e. that act similarly in the simulation and in the reality). 
It allows the algorithm to exploit the simulation and consequently to reduce the number of tests performed on the 
physical robot.

 This approach was successfully used with an E-puck robot in a T-maze and with a quadruped robot that evolved to walk in a straight line with a minimum of transfers on the physical robots \citep{koos2011transferability}. The reality gap phenomenon was particularly apparent in the quadruped experiment: with a controller optimized only in simulation, the virtual robot moved $1.29m$ (in $10s$) but when the same controller was applied on the physical robot, it only moved $0.41m$. With the transferability approach, the obtained solution walked $1.19m$ in simulation and $1.09m$ in reality. These results were found with only $11$ tests on the physical robot and $200,000$ evaluations in simulation. This approach has also been applied for humanoid locomotion \citep{oliveiraoptimization} and damage recovery on a hexapod robot \citep{koos2013fast}.

 Since the transferability approach is one of the most practical tool to apply stochastic optimization algorithms on physical robots, it constitutes an element of our method.

%
%
%
%
%
%
%

\subsection{Novelty Search with Local Competition}
A longstanding challenge in artificial life is to craft an algorithm able to discover a wide diversity of 
interesting artificial creatures. While evolutionary
algorithms are good candidates, they usually converge to a single species of creatures. 
In order to overcome this issue, Lehman and Stanley recently proposed a method called 
\emph{Novelty search with local competition}~\citep{lehman2011evolving}.
This method, based on multi-objective evolutionary algorithms, combines the exploration 
abilities of the Novelty Search algorithm~\citep{lehman2011abandoning} with a
performance competition between similar individuals.

The Novelty Search with Local Competition simultaneously optimizes two objectives for an
individual $\mathbf{c}$: (1) the novelty objective ($novelty(\mathbf{c})$), which measures how 
novel is the individual compared to previously encountered ones, and
(2) the local competition objective ($Qrank(\mathbf{c})$), which compares the individual's quality ($quality(\mathbf{c})$)
to the performance of individuals in a neighborhood, defined with a morphological distance.

With these two objectives, the algorithm favors individuals that are new, 
those that are more efficient than their neighbors and those that are optimal 
trade-offs between novelty and ``local quality''. 
Both objectives are evaluated thanks to an \emph{archive}, which records
all encountered family of individuals and allows the algorithm to define neighborhoods for each individual. 
The novelty objective is computed as the average
distance between the current individual and its neighbors, and 
the local competition objective is the number of neighbors that $\mathbf{c}$ 
outperforms according to the quality criterion $quality(\mathbf{i})$.

The authors successfully applied this method to generate a high number
of creatures with different morphologies, all able to walk in
a straight line. The algorithm found a heterogeneous population of different creatures, from
little hoppers to imposing quadrupeds, all walking at different speeds
according to their stature. We will show in this paper how this algorithm can be modified to allow a single robot to achieve several different actions (i.e. directions of locomotion).



\section{\algoname{}}
\subsection{Main ideas}
Some complex problems are easier to solve when they are split into several sub-problems.
Thus, instead of using a single and complex solution, it is relevant to 
search for several simple solutions that solve a part of the problem. 
This principle is often successfully applied in machine learning: mixtures of experts~\citep{jacobs1991adaptive} 
or boosting~\citep{Schapire1990} methods train several weak classifiers on different 
sub-parts of a problem. Performances of the resulting set of classifiers are 
better than those of a single classifier trained on the whole problem.

The \algoname{} algorithm enables the application of this principle to robotics 
and, particularly, to legged robots that learn to walk.
Instead of learning a complex, inputs-driven controller that generates gaits for every direction, 
we consider a repertoire of un-driven controllers, where each controller is able to reach a different point 
of the space around the robot. This repertoire gathers a high number of efficient and easy-to-learn 
controllers.


Because of the required time, independently learning dozens of controllers is prohibitively expensive, especially with a physical robot. 
To avoid this issue, the \algoname{} algorithm transforms the problem of learning a repertoire
of controllers into a problem of evolving a heterogeneous population of controllers. 
Thus the problem can be solved with an algorithm derived from 
novelty search with local competition~\citep{lehman2011evolving}:
instead of generating virtual creatures with various morphologies that execute the same action,
\algoname{} generates a repertoire of controller, each executing a different action, working on the same creature.
By simultaneously learning all the controllers without the discrimination of a specified goal, 
the algorithm recycles interesting controllers, which are typically wasted with classical 
learning methods. This enhances its optimizing abilities compared to classic optimization methods. 

Furthermore, our algorithm incorporates the transferability approach~\citep{koos2011transferability} to 
reduce the number of tests on the physical robot during the evolutionary process. 
The transferability approach and novelty search with local competition can be combined because 
they are both based on multi-objective optimization algorithms.
By combining these two approaches, the behavioral repertoire is generated in simulation with a virtual robot and only a few controller executions are performed on the physical robot. These trials guide the evolutionary process to solutions that work similarly in simulation and in reality \citep{koos2013fast, koos2011transferability}. 

The minimization of the number of tests on the physical robot and the simultaneous evolution of many controllers are the two assets that allow the \algoname{} algorithm to require significantly less time than classical methods.

More technically, the \algoname{} algorithm relies on four principles, detailed in the next sections:
\begin{itemize}
\item a stochastic, black box, multi-objective optimization 
algorithm simultaneously optimizes 3 objectives, all evaluated in simulation:
(1) the novelty of the gait, (2) the local rank of quality and (3) the local rank of estimated transferability:
\begin{displaymath}
  \textrm{maximize } \left\{\begin{array}{l}
  \textrm{Novelty}(\mathbf{c})\\
 -\textrm{Qrank}(\mathbf{c})\\
 -\widehat{\textrm{Trank}}(\mathbf{c})\\
  \end{array}
  \right.
\end{displaymath}
\item the transferability function is periodically updated with a  test on the physical robot;
\item when a controller is novel enough, it is saved in the \emph{novelty archive};
\item when a controller has a better quality than the one in the archive that reaches the same endpoint, it substitutes the one in the archive.
\end{itemize}

Algorithm \ref{algo} describes the whole algorithm in pseudo-code.





\subsection{Objectives}
The novelty objective fosters the exploration of the reachable space. A controller
is deemed as novel when the controlled individual reaches a region where none, or few of the previously encountered gaits
were able to go (starting from the same point). The novelty score of a controller $\mathbf{c}$ ($Novelty(\mathbf{c})$)
is set as the average distance between the endpoint of the current controller ($\mathcal{E}_c$) 
and the endpoints of controllers contained in its neighborhood ($\mathcal{N}(\mathbf{c})$):
\begin{equation}\label{eq:novelty}
\begin{array}{l}
Novelty(\mathbf{c})=\frac{\sum_{\mathbf{j}\in \mathcal{N}(\mathbf{c})} \|\mathcal{E}_{simu}(\mathbf{c})-\mathcal{E}_{simu}(\mathbf{j})\|}{card(\mathcal{N}(\mathbf{c}))}\\
\end{array}
\end{equation}
To get high novelty scores, individuals have to follow  trajectories leading to endpoints far from the rest of the population.
The population will thus explore all the area reachable by the robot. 
Each time a controller with a novelty score exceeds a threshold ($\rho$), this controller
is saved in an \emph{archive}.
Given this archive and the current population of candidate solutions, a neighborhood is 
defined for each controller ($\mathcal{N}(\mathbf{c})$).
This neighborhood regroups the $k$ controllers that arrive closest to the controller $\mathbf{c}$ (the parameters' values are detailed in the appendix).

The local quality rank promotes controllers that show particular properties, like stability or accuracy.
These properties are evaluated by the quality score ($quality(\mathbf{c})$), which depends on implementation choices and
particularly on the type of  controllers used (we will detail its implementation in section~\ref{sec:implementation}).
In other words, among several controllers that reach the same point, the quality score defines which one should be promoted.
For a controller $\mathbf{c}$, the rank ($Qrank(\mathbf{c})$) is defined as the number of controllers from
its neighborhood that outperform its quality score: minimizing this objective allows the algorithm to find controllers with better quality than their neighbors.
\begin{equation}\label{eq:local}
\begin{array}{l}
Qrank(\mathbf{c})= card(\mathbf{j}\in \mathcal{N}(\mathbf{c}), quality(\mathbf{c})<quality(\mathbf{j}))
\end{array}
\end{equation}

The local transferability rank ($\widehat{Trank(\mathbf{c})}$, equation~\ref{eq:transferability}) works 
as a second local competition objective, where the 
estimation of the transferability score ($\hat{\mathcal{T}}(\mathbf{des(c)})$) replaces the quality score.
Like in~\citep{koos2011transferability}, this estimation  is obtained by periodically repeating three steps:
(1) a controller is randomly selected in the current population or in the archive and then downloaded and executed on the physical robot,
(2) the displacement of the robot is estimated thanks to an embedded sensor, and 
(3) the distance between the endpoint reached in reality and the one in simulation is used to feed a regression model
($\hat\mathcal{T}$, here a support vector machine \citep{chang2011libsvm}). This distance defines the transferability score of the controller. 
This model maps a behavioral descriptor of a controller ($\mathbf{des(c)}$), which is obtained in simulation, with
an approximation of the transferability score. 

Thanks to this descriptor, the regression model predicts the transferability score of each controller in the population and in the archive. 

\begin{equation}\label{eq:transferability}
\begin{array}{l}
\widehat{\textrm{Trank}}(\mathbf{c})= card(\mathbf{j}\in \mathcal{N}(\mathbf{c}), \hat{\mathcal{T}}(\mathbf{des(c)})<\hat{\mathcal{T}}(\mathbf{des(j)}))
\end{array}
\end{equation}

\subsection{Archive management}
In the original novelty search with local competition~\citep{lehman2011evolving}, the archive aims
at recording all encountered solutions, but only the 
first individuals that have a new morphology are added to the archive.
The next individuals with the same morphology are not saved, even if they have better
performances.
In the \algoname{} algorithm, the novelty archive represents the resulting repertoire of controllers,
and thus has to gather only the best controllers for each region of the reachable space.

For this purpose, the archive is differently managed than in the novelty search:
during the learning process, if a controller of the population has better scores ($quality(\mathbf{c})$ or $\hat\mathcal{T}(\mathbf{c})$)
than the closest controller in the archive, the one in the archive is replaced by the better one. 
These comparisons are made with a priority among the scores to prevent circular permutations.
If the transferability score is lower than a threshold ($\tau$), only the transferability
scores are compared, otherwise we compare the quality scores.
This mechanism allows the algorithm to focus the search on transferable controllers instead of searching efficient, but not
transferable, solutions. 
Such a priority is important, as the performances of non-transferable controllers may not be reproducible
on the physical robot.

%
%
%
%
%

\begin{algorithm*}
\small
\caption{\algoname{} algorithm  ( $G$ generations, $T$ transfers' period)}
\label{algo}
\noindent\colorbox[rgb]{0.8,0.8,0.8}{\begin{minipage}{\textwidth}
\begin{algorithmic}
\Procedure{\algoname{}}{}
\State $pop \leftarrow \{c^1, c^2, \ldots, c^{S}\}$ (randomly generated)
\State $archive \leftarrow \emptyset$
\vspace{5pt}
\For{$ g = 1\to G$}
\vspace{5pt}
\ForAll{controller $\mathbf{c} \in pop$}
\vspace{5pt}
\State Execution of $\mathbf{c}$ in simulation
\EndFor
\vspace{5pt}
\If{$g\equiv0[T] $}
\vspace{5pt}
\State\Call{Transferability Update}{$\mathbf{c}$}
\vspace{5pt}
\EndIf

\ForAll{controller $\mathbf{c} \in pop$}
\vspace{5pt}
\State\Call{ Objective Update}{$\mathbf{c}$}
\State\Call{ Archive Management}{$\mathbf{c}$}
\vspace{5pt}
\EndFor

\State Iteration of NSGA-II on $pop$
\EndFor
\State \Return $archive$
\EndProcedure
\end{algorithmic}
\end{minipage}}%

\vspace{5pt}
\noindent\begin{minipage}{\textwidth}
\begin{algorithmic}
\Procedure{Transferability Update}{$\mathbf{c}$}
\State Random selection of $\mathbf{c^*}{\in pop \cup archive}$ and  transfer on the robot
\State Estimation of the endpoint $\mathcal{E}_{real}(\mathbf{c^*})$ 
\State Estimation of the exact transferability value $\Big|\mathcal{E}_{simu}(\mathbf{c^*}) - \mathcal{E}_{real}(\mathbf{c^*})\Big|$
\State Update of the approximated transferability function $\hat{\mathcal{T}}$
\EndProcedure
\vspace{5pt}
\Procedure{Objectives Update}{$\mathbf{c}$}
\State $\mathcal{N}(\mathbf{c}) \leftarrow$ The $15$ controllers($\in pop \cup archive$) closest to $\mathcal{E}_{simu}(\mathbf{c})$
\State Computation of the novelty objective:
\State \hspace{15pt} $Novelty(\mathbf{c})=\frac{\sum_{\mathbf{j}\in \mathcal{N}(\mathbf{c})} \|\mathcal{E}_{simu}(\mathbf{c})-\mathcal{E}_{simu}(\mathbf{j})\|}{|\mathcal{N}(\mathbf{c})|}$
\State Computation of the local rank objectives:
\State \hspace{15pt}$Qrank(\mathbf{c})= |\mathbf{j}\in \mathcal{N}(\mathbf{c}), quality(\mathbf{c})<quality(\mathbf{j})|$
\State \hspace{15pt}$\widehat{\textrm{Trank}}(\mathbf{c})= |\mathbf{j}\in \mathcal{N}(\mathbf{c}), \hat{\mathcal{T}}(\mathbf{des(c)})<\hat{\mathcal{T}}(\mathbf{des(j)}|$
\EndProcedure
\vspace{5pt}
\Procedure{Archive Management}{$\mathbf{c}$}
\If{$Novelty(\mathbf{c})> \rho$}
\State Add the individual to $archive$
\EndIf
$\mathbf{c}_{nearest} \leftarrow$ The controller $\in archive$ nearest to $\mathcal{E}_{simu}(\mathbf{c})$
\If{ $\hat{\mathcal{T}}(\mathbf{des(c)})> \tau$  and  $quality(\mathbf{c}) > quality(\mathbf{c}_{nearest}) $}
\State Replace $\mathbf{c}_{nearest}$ by $\mathbf{c}$ in the archive
\ElsIf{$\hat{\mathcal{T}}(\mathbf{des(c)})>\hat{\mathcal{T}}(\mathbf{des(c}_{nearest}))$}
\State Replace $\mathbf{c}_{nearest}$ by $\mathbf{c}$ in the archive
\EndIf
\EndProcedure
\end{algorithmic}
\end{minipage}
\end{algorithm*}

\section{Experimental Validation}\label{section:experiment}

\begin{figure}
\includegraphics[width=\linewidth]{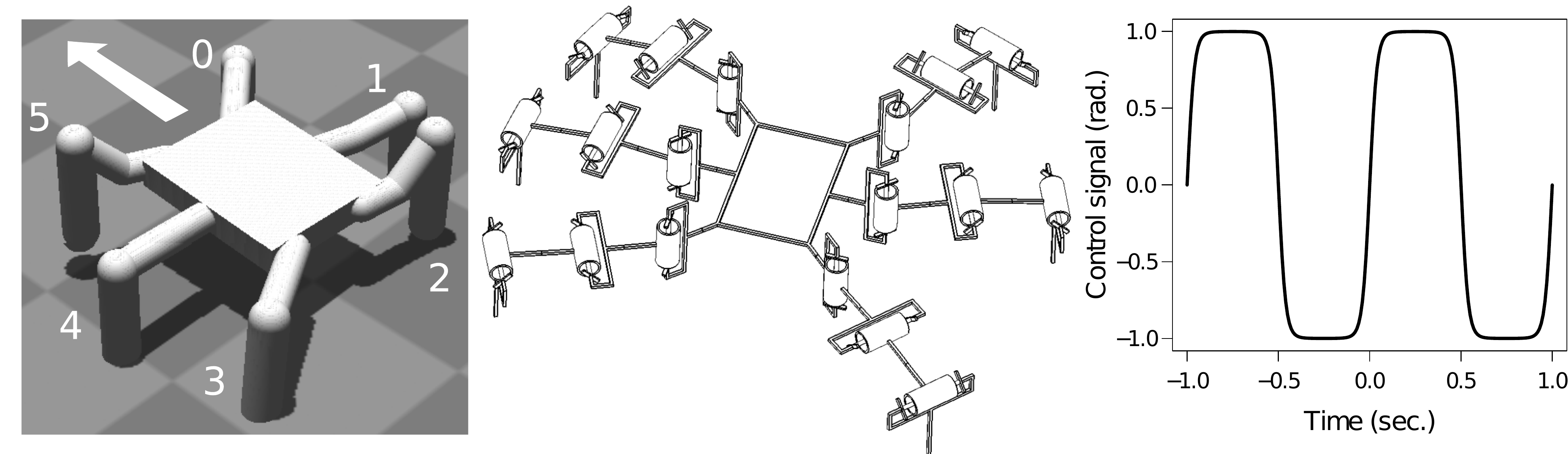}
\caption{(Left) Snapshot of the simulated robot in our ODE-based
  physics simulator. (Center) Kinematic scheme of the robot. The cylinders
  represent actuated pivot joints. (Right) Control function
  $\gamma(t,\ \alpha,\ \phi)$ with $\alpha =1$ and $\phi=0$.}
\label{fig:virtual_robot}\label{fig:cinematique}
\end{figure}

We evaluate the \algoname{} algorithm on two different experiments, which both consist in evolving a repertoire of controllers to access to the whole vicinity of the robot. In the first experiment, the algorithm is applied on a simulated robot (Fig. \ref{fig:virtual_robot}, left), consequently the transferability aspect of the algorithm is disabled. The goal of this experiment is to show the benefits of evolving simultaneously all the behaviors of the repertoire instead of evolving them separately. The second experiment applies the algorithm directly on a physical robot (Fig \ref{fig:objective}, left). For this experiment, the transferability aspect of the algorithm is enabled and the experiment shows how the behavioral repertoire can be learned with a few trials on the physical robot.

\subsection{Implementation choices}\label{sec:implementation}

The pseudo-code of the algorithm is presented in Algorithm \ref{algo}. The \algoname{} algorithm uses the same variant of NSGA-II \citep{deb2002fast} as the novelty search with local competition \citep{lehman2011evolving}.  
The simulation of the robot is based on the \emph{Open Dynamic Engine} (ODE) and the transferability function $\hat{\mathcal{T}}$ uses the $\nu$-Support Vector Regression algorithm with linear kernels implemented in the library \emph{libsvm}~\citep{chang2011libsvm} (learning parameters set to default values).  All the algorithms are implemented in the Sferes$_{v2}$ framework~\citep{Mouret2010} (parameters and source code are detailed in appendix). The simulated parts of the algorithms are computed on a cluster of 5 quad-core Xeon-E5520@2.27GHz computers.

\subsubsection{Robot}

Both the virtual and the physical robots have the same kinematic scheme (see figure \ref{fig:cinematique} center). They have 18 degrees of freedom, 3 per leg. The first joint of each leg controls the direction of the leg while the two others define its elevation and extension.  The virtual robot is designed to be a ``virtual copy'' of the physical hexapod: it has the same mass for each of its body parts, and the physical simulator reproduces the dynamical characteristics of the servos. On the physical robot, the estimations of the covered distance are acquired with a Simultaneous Localisation and Mapping (SLAM) algorithm based on the embedded RGB-D sensor \citep{endres12icra}.

\subsubsection{Genotype and Controller}
The same genotype and controller structure are used for the two
  sets of experiments. The genotype is a set of 24 parameter values
  defining the angular position of each leg joint with a periodic
  function $\gamma$ of time $t$, parametrized by an amplitude $\alpha$
  and a phase shift $\phi$ (Fig. \ref{fig:virtual_robot}, right):


\begin{equation}
\gamma(t,\ \alpha,\ \phi) = \alpha\cdot\tanh\left(4\cdot\sin\left(2\cdot\pi\cdot(t + \phi)\right)\right)\label{eq:tanh}
\end{equation}

Angular positions are updated and sent to the servos every 30ms. The
main feature of this particular function is that the control signal is
constant during a large portion of each cycle, thus allowing the robot
to stabilize itself.  In order to keep the ``tibia'' of each leg
vertical, the control signal of the third servo is the opposite of the second one. 
Consequently, positions sent to the $i^{th}$ leg are:
\begin{itemize}
\item $\gamma(t,\ \alpha^i_1,\ \phi^i_1)$ for servo 1;
\item $\gamma(t,\ \alpha^i_2,\ \phi^i_2)$ for servos 2;
\item $-\gamma(t,\ \alpha^i_2,\ \phi^i_2)$ for servos 3.
\end{itemize}

The $24$ parameters can each have five different values ($0, 0.25, 0.5,
0.75, 1$) and with their variations, numerous gaits are possible, from
purely quadruped gaits to classic tripod gaits. 

For the genotype mutation, each parameter value has a 10\% chance of being changed to any value in the set of possible values, with the new value chosen randomly from a uniform distribution over the possible values. For both of the experiments, the crossover is disabled.

Compared to classic evolutionary algorithms, \algoname{} only changes the way individuals are selected. As a result, it does not put any constraint on the type of controllers, and many other controllers are conceivable (e.g. bio-inspired central pattern generators \citep{sproewitz2008learning,Ijspeert2008}, dynamic movement primitives~\citep{schaal2003dynamic} or evolved neural networks~\citep{Yosinski2011,Clune2011}).

\subsubsection{Endpoints of a controller}
The endpoint of a controller (in simulation or in reality) is the position of the center of the
robot's body projected in the horizontal plane after running the controller for
$3$ seconds:
\begin{displaymath}
\mathcal{E}(\mathbf{c})=\left\{\begin{array}{l}
  \textrm{center}_x(t=3s)-\textrm{center}_x(t=0s)\\
  \textrm{center}_y(t=3s)-\textrm{center}_y(t=0s)
  \end{array}
  \right\}
\end{displaymath}

\subsubsection{Quality Score}

To be able to sequentially execute saved behaviors, special attention is paid to the final orientation of the robot. Because the endpoint of a trajectory depends on the initial orientation of the robot, we need to know  how the robot ends its previous movement when we plan the next one. To facilitate chaining controllers, we encourage behaviors that end their movements with an orientation aligned with their trajectory.

\begin{figure}
\includegraphics[width=\linewidth]{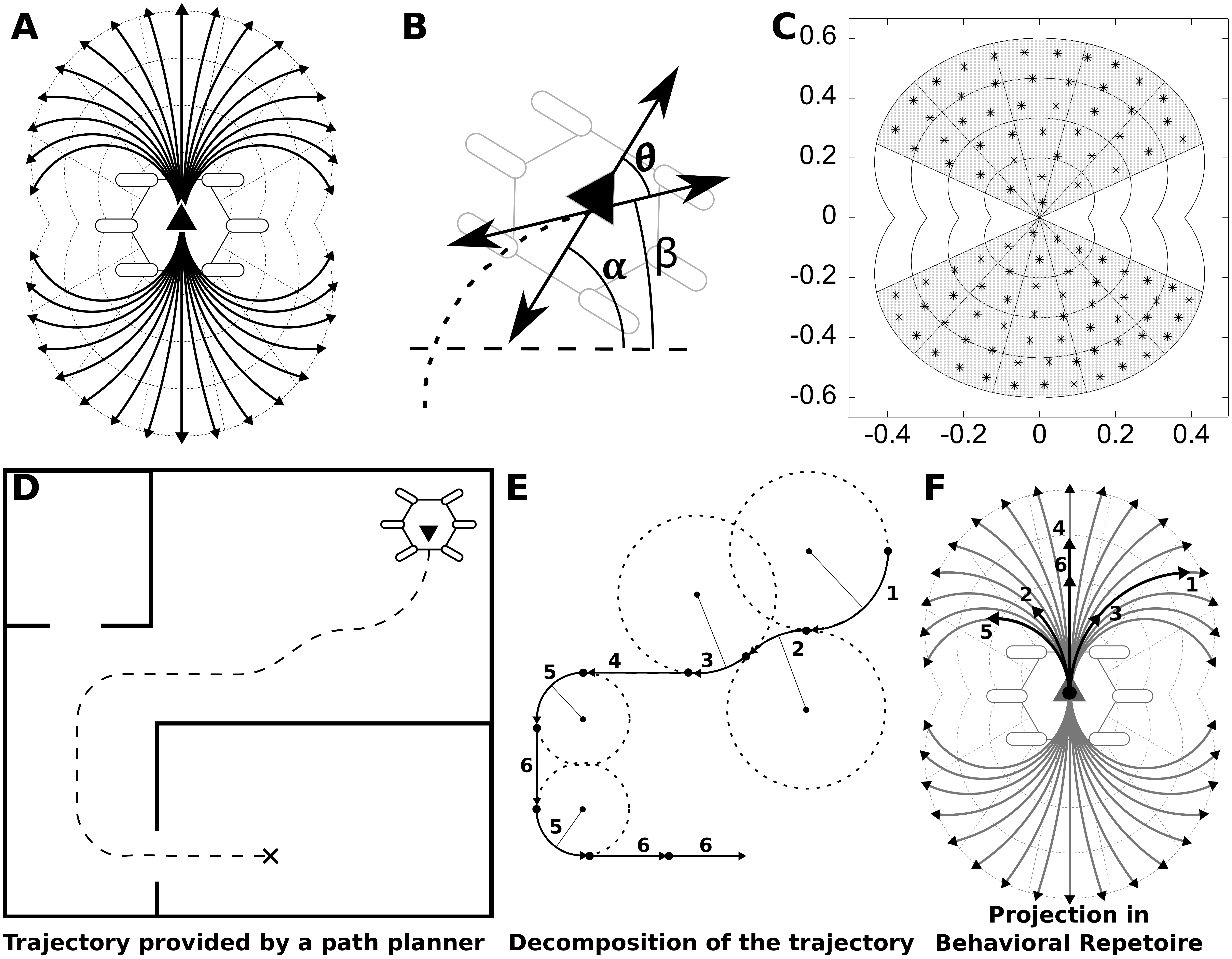}
\caption{(A) Examples of trajectories following a circle centered on
  the lateral axis with several radii. (B) Definition of the desired
  orientation. $\theta$ represents the orientation error between
  $\alpha$, the final orientation of the robot, and $\beta$, the
  tangent of the desired trajectory. These angles are defined
  according to the actual endpoint of the individual, not the desired
  one.  (C) Reachable area of the robot viewed from top.  A region of
  interest (ROI) is defined to facilitate post-hoc analysis (gray
  zone). The boundaries of the region are defined by two lines at 60 degrees on
  each side of the robot. The curved frontier is made of all the reachable
  points with a curvi-linear abscissa lower than 0.6 meters (
  these values were set thanks to experimental observations of
  commonly reachable points). Dots correspond to targets selected for
  the control experiments. (D-E-F) Illustration of how
    a behavioral repertoire can be used with a hexapod robot. First, a path planning algorithm
    computes a trajectory made of lines and portions of circles \citep{lavalle2006planning,Siciliano2008}. Second, to follow this trajectory, the robot sequentially executes the most appropriate behavior in the repertoire (here numbered on E and F). For closed-loop control, the trajectory can be re-computed at each time-step using the actual position of the robot.}
\label{fig:circle_traj}
\end{figure}

The robot cannot execute arbitrary trajectories with a single controller because controllers are made of simple periodic functions. For example, it cannot begin its movement by a turn and then go straight. With this controller, the robot can only follow trajectories with a constant curvature, but it can still can move sideways, or even turn around itself while following an overall straight trajectory. We chose to focus the search on circular trajectories, centered on the lateral axis, with a variable radius (Fig.~\ref{fig:circle_traj}A), and for which the robot's body is pointing towards the tangent of the overall trajectory. Straight, forward (or backward) trajectories are still possible with the particular case of an infinite radius. This kind of trajectory is suitable for motion control as many complex trajectories can be decomposed in a succession of circle portions and lines.
An illustration of this principle is pictured on figure~\ref{fig:circle_traj} (D-E-F).

To encourage the population to follow these trajectories, the quality
score is set as the angular difference between the arrival orientation and 
the tangent of the circular trajectory that corresponds to the endpoint (Fig.~\ref{fig:circle_traj}B):
\begin{equation}
quality(\mathbf{c})= - |\theta(\mathbf{c})|= - |\alpha(\mathbf{c})-\beta(\mathbf{c})|
\end{equation}


\subsubsection{Transferability score}
The transferability score of a tested controller $\mathbf{c^*}$ is computed as the distance between the controller's endpoint reached in simulation and the one reached in reality:
\begin{equation}
\textrm{transferability}(\mathbf{c^*})= - |\mathcal{E}_{\mathrm{simu}}-\mathcal{E}_{\mathrm{real}}|
\end{equation}

In order to estimate the transferability score of untested controllers,
a regression model is trained with the tested controllers and their
recorded transferability score.  The regression model used is the
$\nu$-Support Vector Regression algorithm with linear kernels
implemented in the library \emph{libsvm}~\citep{chang2011libsvm}
(learning parameters are set to default values), which maps a
behavioral descriptor ($\mathbf{des(c)}$) with an estimated transferability score ($\mathcal{T}(\mathbf{des(c)})$).  Each
controller is described with a vector of Boolean values that describe,
for each time-step and each leg, whether the leg is in contact with
the ground (the descriptor is therefore a vector of size $N\times6$,
where $N$ is the number of time-steps). This kind of vector is a
classic way to describe gaits in legged animals and
robots~\citep{raibert1986legged}.  
During the evolutionary process, the algorithm performs $1$ transfer every $50$ iterations.

\subsection{Experiments on the Virtual Robot} \label{sec:virtual_results}
This first experiment involves a virtual robot that learns a
behavioral repertoire to reach every point in its vicinity. The
transferability objective is disabled because the goal of this
experiment is to show the benefits of learning simultaneously all the
behaviors of the repertoire instead of learning them separately. Using
only the simulation allows us to perform more replications and to
implement a higher number of control experiments. This experiment also
shows how the robot is able to \emph{autonomously}:
\begin{itemize}
\item discover possible movements;
\item cover a high proportion of the reachable space;
\item generate a behavioral repertoire.
\end{itemize}

The \algoname{} experiment and the control experiments (described in the next section) are replicated
40 times to gather statistics.

\subsubsection{Control Experiments}
To our knowledge, no work directly tackles the question of learning simultaneously all the behaviors of a controller repertoire,
thus we cannot compare our approach with an existing method.
As a reference point, we implemented a naive method where the desired endpoints 
are preselected. A different controller will be optimized to reach each different wanted point individually. 

We define 100 target points, spread thanks to a K-means algorithm \cite{seber1984multivariate}  over the defined region of interest (ROI) of the reachable area (see Fig. \ref{fig:circle_traj}C).
We then execute several multi-objective evolutionary algorithms (NSGA-II \cite{deb2002fast}), one for each reference point. 
At the end of each execution of the algorithm, the nearest individual to the target point in the Pareto-front is saved in an archive.
This experiment is called ``nearest'' variant.
We also save the controller with the best orientation (quality score described previously) within a radius of 10 cm around the target point
and we call this variant ``orientation''.
The objectives used for the optimization are:
\begin{displaymath}
  \textrm{minimize } \left\{\begin{array}{l}
  Distance(\mathbf{c})=\|E_\mathbf{c}-E_{Reference}\|\\
  Orientation(\mathbf{c})=|\alpha(\mathbf{c})-\beta(\mathbf{c})|
  \end{array}
  \right.
\end{displaymath}

We also investigate how the archive management added in
\algoname{} improves the quality of produced behavioral repertoires.
To highlight these improvements, we compared our resulting archives with archives issued from the Novelty Search algorithm \citep{lehman2011abandoning} 
and from the Novelty Search with Local
Competition algorithm \citep{lehman2011evolving}, as the main difference between these algorithms is archive management procedure.
We apply these algorithms on the
same task and with the same parameters as in the experiment with our
method. We call these experiments ``Novelty Search''(NS) and ``NS with Local Competition''. 
For both of these experiments we will analyze both the produced archives and the resulting populations.

For all the experiments we will study the sparseness and the orientation error
of the behavioral repertoires generated by each approach.  All these
measures are done within the region of interest previously defined. The
sparseness of the archive is computed by discretizing the ROI with a
one centimeter grid ($\mathcal{G}$), and for each point $p$ of that grid the distance from
the nearest individual of the archive ($\mathcal{A}$) is recorded.  The sparseness of
the archive is the average of all the recorded distances:
\begin{equation}
 \textrm{sparseness}(\mathcal{A})=\frac{ \sum_{p \in \mathcal{G}}^{} \min_{i \in \mathcal{A}} (\textrm{distance}(i,p))}{\textrm{card}(\mathcal{G})}
\end{equation}
where $\textrm{card}(\mathcal{G})$ denotes the number of elements in $\mathcal{G}$.

The quality of the archive is defined as the average orientation error for all the individuals inside the ROI:
\begin{equation}
\textrm{Orientation Error}(\mathcal{A})=\frac{\sum_{i\in\mathcal{A}\in ROI} \theta(i) }{  \textrm{card}(\mathcal{A}\in\textrm{ROI})}
\end{equation}

\subsubsection{Results}
\begin{figure*}
\centering
\includegraphics[width=\linewidth]{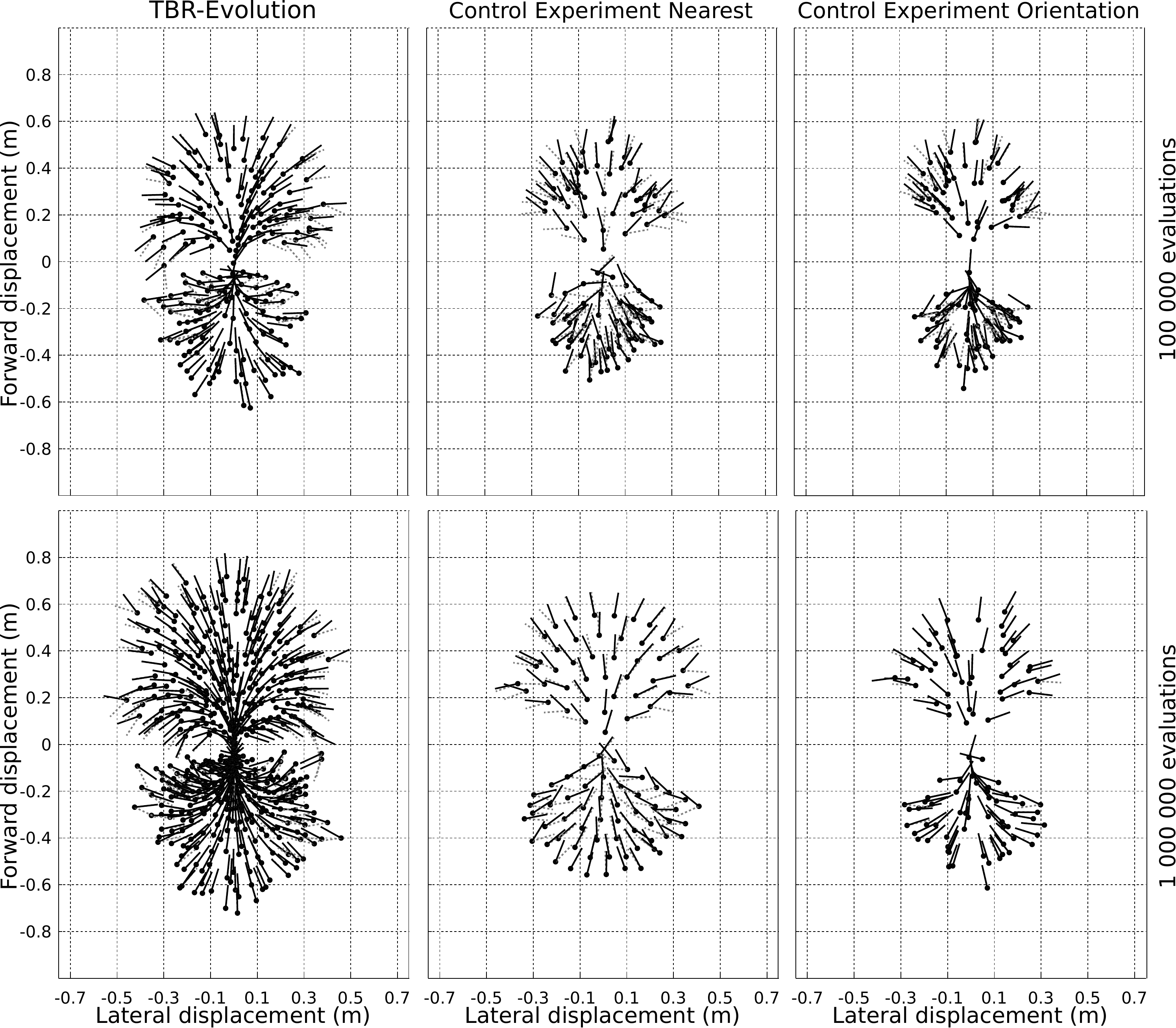}
\caption{Comparison between the typical results of the \algoname{} algorithm, the ``nearest'', and the ``orientation''. The archives are displayed
after 100 000 evaluations (top) and after 1 000 000 evaluations (bottom).
Each dot corresponds to the endpoint of a controller.
The solid lines represent the final orientation of the robot for each controller, while the gray dashed lines represent the desired orientation.
The orientation error is the angle between solid and dashed lines.}
\label{fig:archive_example}
\end{figure*}

\begin{figure*}
\centering
\includegraphics[width=\linewidth]{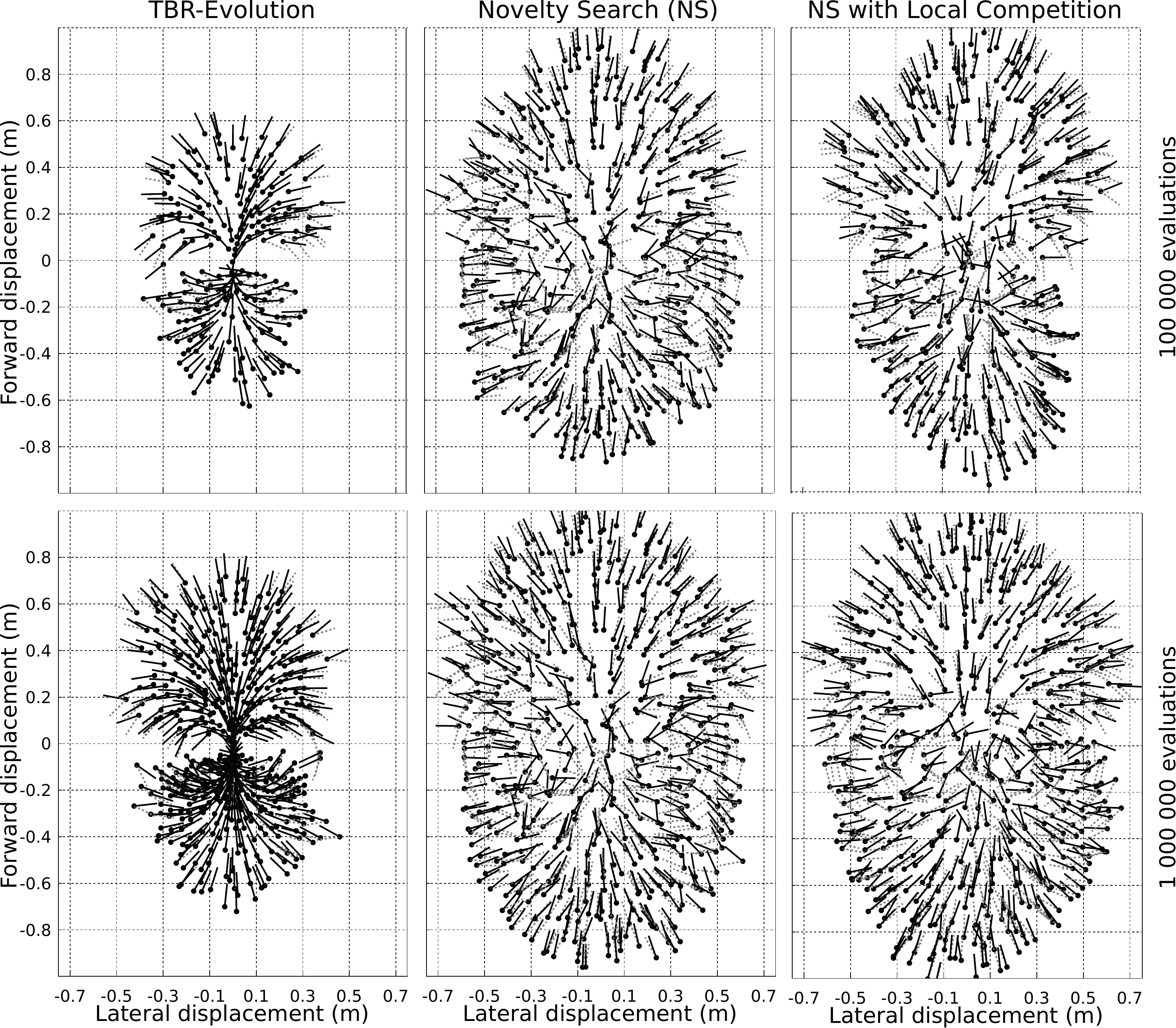}
\caption{Comparison between the typical results of the \algoname{} algorithm, the Novelty Search, and the NS with Local Competition. The archives are displayed
after 100 000 evaluations and after 1 000 000 evaluations.
Each dot corresponds to the endpoint of a controller.
The solid lines represent the final orientation of the robot for each controller, while the gray dashed lines represent the desired orientation.
The orientation error is the angle between solid and dashed lines.}
\label{fig:archive_example_2}
\end{figure*}

\begin{figure*}
\centering
\includegraphics[width=\linewidth]{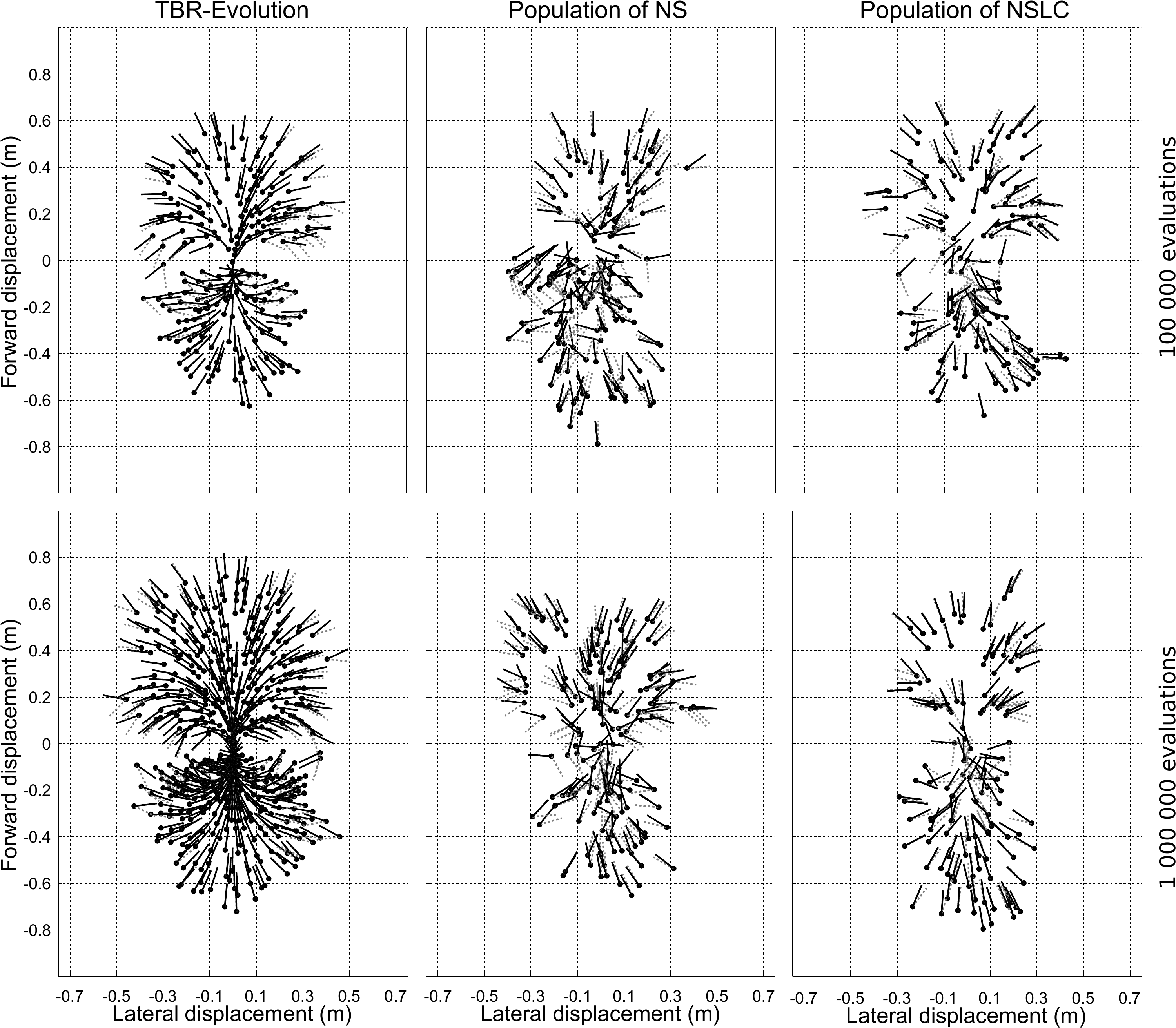}
\caption{Comparison between typical results of the \algoname{} algorithm, the population of  Novelty Search, and the population of NS with Local Competition. The archives/populations are displayed after 100 000 evaluations and after 1 000 000 evaluations.
Each dot corresponds to the endpoint of a controller.
The solid lines represent the final orientation of the robot for each controller, while the gray dashed lines represent the desired orientation.
The orientation error is the angle between solid and dashed lines.}
\label{fig:archive_example_3}
\end{figure*}

\begin{figure*}
\includegraphics[width=\linewidth]{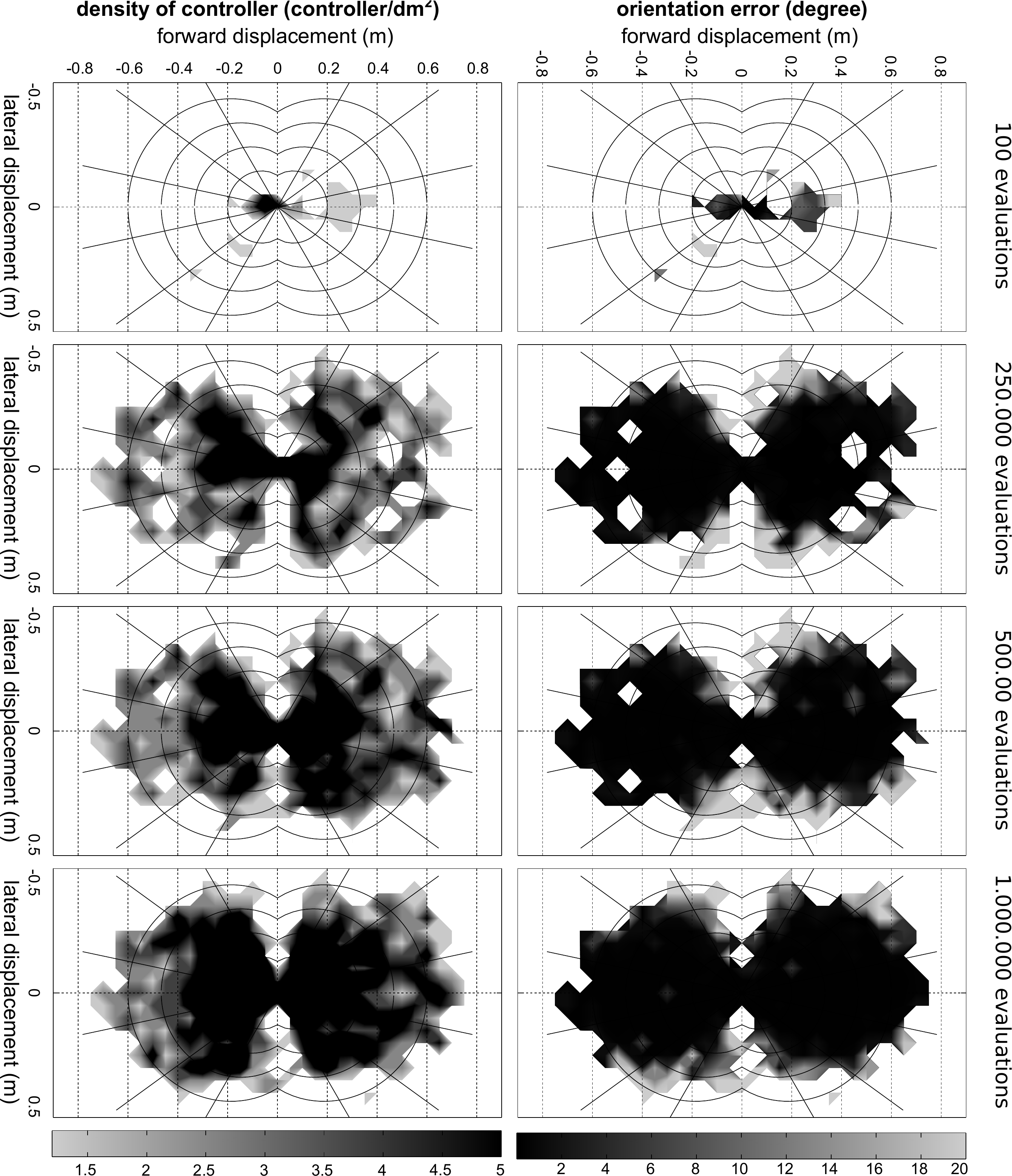}
\caption{(Left) Variation of density of controller (number of controllers per $dm^2$).
 (Right) Variation of the orientation error (given by the nearest controller) along a typical run in simulation.}
\label{fig:evolution}
\end{figure*}

Resulting behavioral repertoires from a typical run of \algoname{} and the 
control experiments are pictured on figures \ref{fig:archive_example}, \ref{fig:archive_example_2} and \ref{fig:evolution}.
The endpoints achieved with each controller of the repertoire are spread over the reachable space in
a specific manner: they cover both the front and the back of the
robot, but less the lateral sides.  These limits are not explicitly
defined, but they are autonomously discovered by the
algorithm.

For the same number of evaluations, the area is less covered with the
control experiments (nearest and orientation) than with \algoname{}
(Fig. \ref{fig:archive_example}). With only 100 000 evaluations, this area
is about twice larger with \algoname{} than with both control
experiments.  At the end of the evolution (1 000 000 evaluations), the
reachable space is more dense with our approach. With the ``nearest''
variant of the control experiment, all target points are reached (see
Fig. \ref{fig:circle_traj}C), this is not the case for the
``orientation'' variant.

The archives produced by Novelty Search and NS with Local Competition
both cover a larger space than the \algoname{} algorithm
(Fig. \ref{fig:archive_example_2}). These results are surprising
because all these experiments are based on novelty search and differ
only in the way the archive is managed. These results show that
\algoname{} tends to slightly reduce the exploration abilities of NS
and focuses more on the quality of the solutions.  

We can formulate two hypotheses to explain this difference in exploration. First, the ``local competition'' objective may have an higher influence in the \algoname{} algorithm than in the NS with Local Competition: in NS with local competition, the individuals from the population are competing against those of the archive; since this archive is not updated if an individual with a similar behavior but a higher performance is encountered, the individuals from the population are likely to always compete against low-performing individuals, and therefore always get a similar local competition score; as a result, the local competition objective is likely to not be very distinctive and most of the selective pressure can be expected to come from the novelty objective. This different selective pressure can explain why NS with local competition explores more than BR-Evolution, and it echoes the observation that the archive obtained with NS and NS with local competition are visually similar (Fig. \ref{fig:archive_example_2}). The second hypothesis is that the procedure used to update the archive may erode the borderline of the archive: if a new individual is located close to the archive's borderline, and if this individual has a better performance than its nearest neighbor in the archive, then the archive management procedure of \algoname{} will replace the individual from the archive with the new and higher-performing one; as a consequence, an individual from the border can be removed in favor of a higher-performing but less innovative individual. This process is likely to repeatedly ``erode'' the border of the archive and hence discourage exploration. These two hypotheses will be investigated in future work.

The primary purpose of the Novelty Search with Local Competition is to maintain a diverse variety of well adapted solutions in its population, and not in its archive. For this reason, we also plotted the distribution of the population's individuals for both the Novelty Search and the NS with Local Competition (Fig. \ref{fig:archive_example_3}). After 100,000 evaluations, and at the end of the evolution, the population covers less of the robot's surrounding than \algoname{}. The density of the individuals is not homogeneous and they are not arranged in a particular shape, contrary to the results of \algoname{}. In particular, the borderline of the population seems to be almost random.

The density of the archive is also different between the algorithms. The density of the archives produced by \algoname{} is higher than the other approaches, while the threshold of novelty ($\rho$) required to add individuals in the archive is the same. This shows that the archive management of the \algoname{} algorithm increases the density of the regions where solutions with a good quality are easier to find. This characteristic allows a better resolution of the archive in specific regions.

The orientation error is qualitatively more important in the
``nearest'' control experiment during all the evolution than with the
other experiments. This error is important at the beginning of the
``orientation'' variant too, but, at the end, the error is negligible
for the majority of controllers.  The Novelty Search, NS with local
competition and the population of the Novelty Search have a
larger orientation error, the figures \ref{fig:archive_example_2}
and \ref{fig:archive_example_3} show that the orientation of
the controllers seems almost random. With such repertoire, chaining
behaviors on the robot is more complicated than with the \algoname{}'s
archives, where a vector field is visible.  Only the population
  of the NS with Local Competition seems to show lower orientation error. This
  illustrates the benefits of the local competition objective on the
  population's behaviors.

The \algoname{} algorithm consistently leads to very small orientation errors (Fig. \ref{fig:archive_example} and Fig. \ref{fig:evolution}); only few points have a significant error. We find these points in two regions, far from the starting point and directly on its sides. These regions are characterized by their difficulty to be accessed, which stems from two main causes: the large distance to the starting point or the complexity of the required trajectory given the controller and the possible parameters (Appendix \ref{sec:implementation}). For example the close lateral regions require executing a trajectory with a very high curvature, which cannot be executed with the range of parameters of the controller. Moreover, the behaviors obtained in these regions are most of the time degenerated: they take advantages of inacurracies in the simulator to realize movement that would not be possible in reality. Since accessing these points is difficult, finding better solutions is difficult for the evolutionary algorithm. We also observe a correlation between the density of controllers, the orientation error and the regions difficult to access (Fig. \ref{fig:evolution}): the more a region is difficult to access, the less we find controllers, and the less these controllers have a good orientation. For the others regions, the algorithm produces behaviors with various lengths and curvatures, covering all the reachable area of the robot.

\begin{figure}
\includegraphics[width=\linewidth]{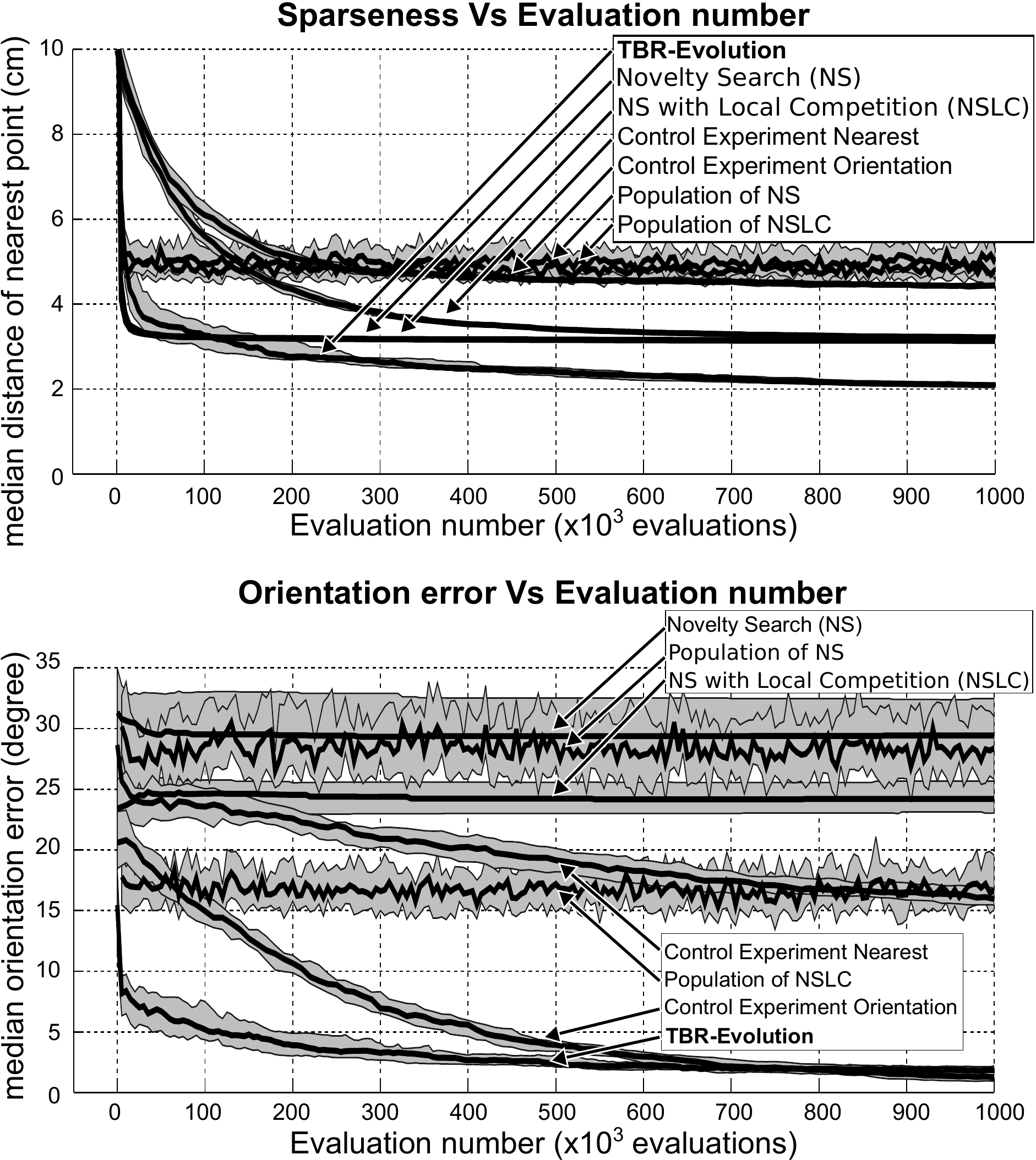}
\caption{(Top) Variation of the sparseness of the controller repertoire. For each point of a one centimeter grid inside the ROI (Fig. \ref{fig:circle_traj}), the distance from the nearest 
controller is computed. The sparseness value is the average of these distances.
This graph plots the first three quartiles of the sparseness computed with 40 
runs for each algorithm. (Bottom) Variation of the median of the orientation error over all the controllers inside the 
region of interest. This graph also plots the three first quartiles ($25\%, 50\%, 75\%$) computed
with 40 runs for each algorithm.}

\label{fig:precision}
\end{figure}

In order to get a statistical point of view, we studied the median,
over 40 runs, of the sparseness and the quality of controllers inside
a region of interest (ROI) (Fig. \ref{fig:precision}, Top).  The
\algoname{} algorithm achieved a low sparseness value with few
evaluations. After 100 000 evaluations, it was able to generate
behaviors covering the reachable space with an interval distance of
about 3 cm.  At the end of the process, the sparseness value is near 2
cm.  With the ``nearest'' and the ``orientation'' experiments, the
variation is slower and reaches a significantly higher level of
sparseness (p-values $=1.4\times10^{-14}$ with Wilcoxon rank-sum
tests). The ``orientation'' variant of the control experiment exhibits
the worst sparseness value ($>4 cm$). This result is expected because
this variant favors behaviors with a good orientation even if they are
far from their reference point. This phenomenon leads to a sample of
the space less evenly distributed. The ``nearest'' variant achieves
every target points, thus the sparseness value is better than with the
``orientation'' variant (3 cm vs 4cm, at the end of the experiment).
The Novelty Search and the NS with Local Competition experiments
follow the same progression (the two lines are indistinguishable) and
reach their final value faster than the \algoname{} algorithm. As our
algorithm can increase the density of controller in particular
regions, at the end of the evolution, the final value of sparseness of
\algoname{} is better than all the control experiments. The
  sparseness of the populations of Novelty Search and NS with Local Competition are indistinguishable too, but also
  constant over all the evolution and larger than all the tested
  algorithms, mainly because of the uneven distribution of their individuals (fig. \ref{fig:archive_example_3})

From the orientation point of view (Fig. \ref{fig:precision}, bottom),
our approach needs few evaluations to reach a low error value (less
than $5$ degrees after 100 000 evaluations and less than $1.7$ degrees
at the end of the evolutionary process).  The variation of the
``orientation'' control experiment is slower and needs 750 000
evaluations to cross the curve of \algoname{}.  At the end of the
experiment this variant reaches a significantly lower error level
(p-values $=3.0\times10^{-7}$ with Wilcoxon rank-sum tests), but this
corresponds to a difference of the medians of only $0.5$ degrees.  The
``nearest'' variant suffers from significantly higher orientation
error (greater than $15$ degrees, p-values $=1.4\times10^{-14}$ with
Wilcoxon rank-sum tests). This is expected because this variant
selects behaviors taking into account only the distance from the
target point. With this selection, the orientation aspect is
neglected. The Novelty Search and the NS with Local Competition
experiments lead to orientation errors that are very high and almost
constant over all the evolution. These results come from the archive
management of these algorithms which do not substitute individuals
when a better one is found. The archive of these algorithms only
gathers the first encountered behavior of each reached point.  The
orientation error of the NS with Local Competition is lower than the
Novelty Search because the local competition promotes behavior with a
good orientation error (compared to their local niche) in the
population, which has an indirect impact on the quality of the archive
but not enough to reach a low error level.  The same conclusion
can be drawn with the population of these two algorithms: while the
populations of the Novelty Search have a similar orientation error
than its archives, the populations of the NS with Local Competition
have a lower orientation error than its archives.

\begin{figure}
\includegraphics[width=\linewidth]{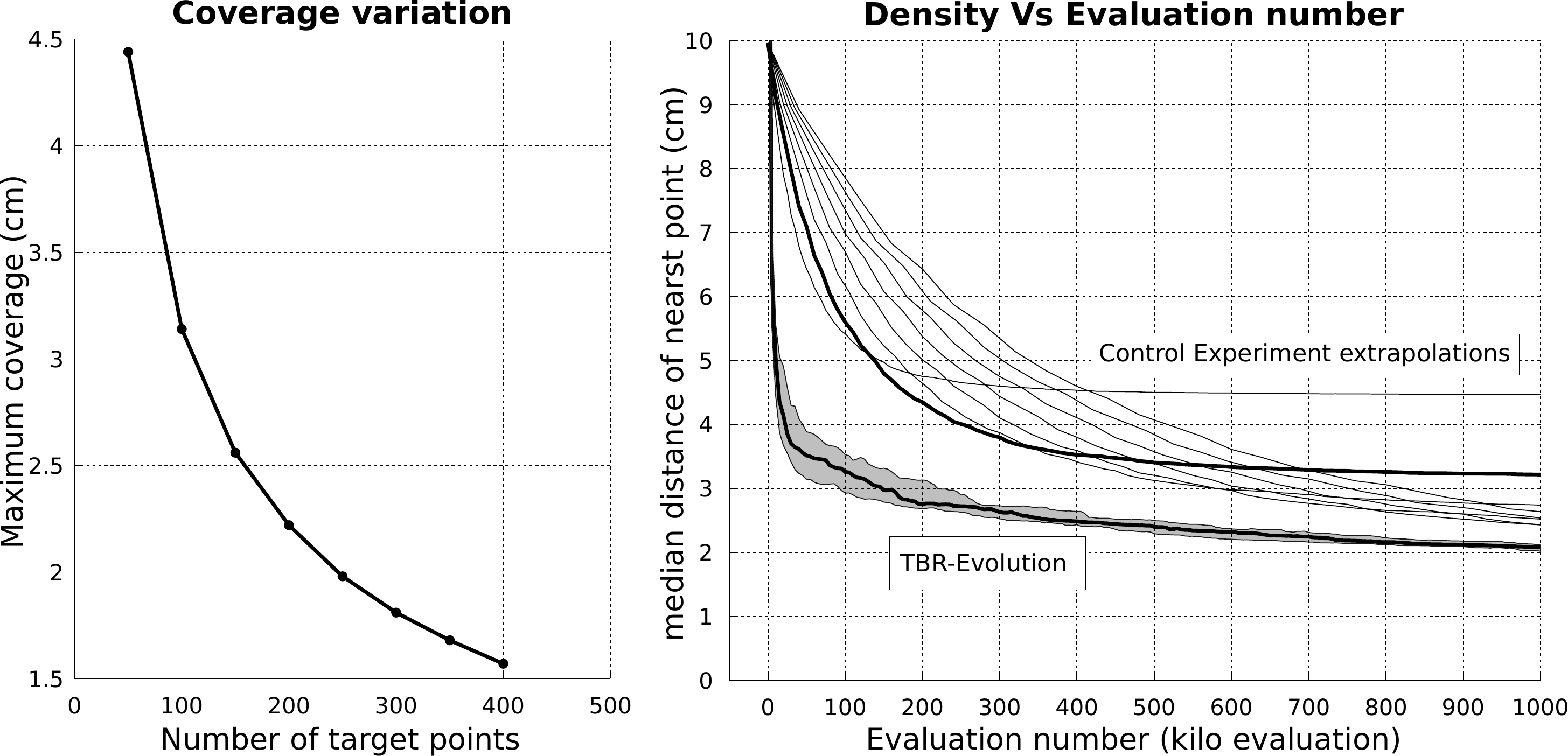}
\caption{(Left) Theoretical sparseness of the control experiments
  according to the number of points. With more points, the sparseness
  value will be better (lower). (Right) Extrapolations of the
  variation of the sparseness for the ``nearest'' variant of the
  control experiment according to different number of targets.  Each
  line is an extrapolation of the variation of the sparseness of the
  ``nearest variant'', which are based on a number of points starting from 50 to
  400, with a 50 points step. The variation of \algoname{} is also
  plotted for comparison.}
\label{fig:extrapolations}
\end{figure}

With the sets of reference points, we can compute the theoretical minimal sparseness value of the control experiments (Fig. \ref{fig:extrapolations}, Left).  For example, changing the number of targets from 100 to 200 will change the sparseness value from 3.14 cm to 2.22 cm.  Nonetheless, doubling the number of points will double the required number of evaluations. Thanks to these values we can extrapolate the variation of the sparseness according to the number of points. For example, with 200 targets, we can predict that the final value of the sparseness will be 2.22 and thus we can scale the graph of our control experiment to fit this prediction.  Increasing the number of targets will necessarily increase the number of evaluations, for example using 200 targets will double the number of evaluations. Following this constraint, we can also scale the temporal axis of our control experiment. We can thus extrapolate the sparseness of the archive with regard o the number of target, and compare it to the sparseness of the archive generated with \algoname{}.

The extrapolations (Fig. \ref{fig:extrapolations}, right) show higher sparseness values compared to \algoname{} within the same execution time. Better values will be achieved with more evaluations. For instance, with 400 targets the sparseness value reaches 1.57 cm, but only after 4 millions of evaluations. This figure shows how our approach is faster than the control experiments regardless the number of reference points.

Figures \ref{fig:precision} and \ref{fig:extrapolations} demonstrate 
how \algoname{} is better both in the sparseness and in the orientation aspects
compared than the control experiments. 
Within few evaluations, reachable points are evenly distributed around the robot 
and corresponding behaviors are mainly well oriented.

(An illustrating video is available on: 
\url{http://youtu.be/2aTIL_c-qwA})

\subsection{Experiments on the Physical Robot}

In this second set of experiments, we apply the \algoname{} algorithm on a physical hexapod robot (see Fig. \ref{fig:objective} left). The transferability component of the algorithm allows it to evolve the behavioral repertoire with a minimum of evaluation on the physical robot. For this experiment, $3 000$ generations are performed and we execute a transfer (evaluation of one controller on the physical robot) every 50 generations, leading to a total of $60$ transfers. The \algoname{} experiments and the reference experiments are replicated $5$ times to gather statistics\footnote{Performing statistical analysis with only 5 runs is difficult but it still allows us to understand the main tendencies. The current set of experiments (5 runs of TBR-Evolution and the control experiment) requires more than 30 hours with the robot and it is materially challenging to use more replications.}.

\subsubsection{Reference Experiment}
In order to compare the learning speed of the \algoname{} algorithm, we use a reference experiment where only one controller is learned. For this experiment, we use the NSGA-II algorithm with the transferability approach to learn a controller that reaches a predefined target. The target is situated 0.4m in front and 0.3m to the right: a point not as easy to be accessed as going only straight forward, and not as hard as executing a U-turn. It represents a good difficulty trade-off and thus allows us to extrapolate the performances of this method to more points.

The main objective is the distance ($\textrm{Distance}(\mathbf{c})$) between the endpoint of the considered controller
and the target. The algorithm also optimizes the estimated transferability value ($\hat\mathcal{T}(\mathbf{des(c)})$) and the orientation
error ($\textrm{perf}(\mathbf{c})$) with the same definition as in the \algoname{} algorithm:
\begin{displaymath}
  \textrm{minimize } \left\{\begin{array}{l}
  \textrm{Distance}(\mathbf{c})\\
  \hat\mathcal{T}(\mathbf{des(c)})\\
  \textrm{perf}(\mathbf{c})\\
  \end{array}
  \right.
\end{displaymath}

To update the transferability function, we use the same transfer
frequency as in \algoname{} experiments (every 50 generations).  Among
the resulting trade-offs, we select as final controller the one that
arrives closest to the target among those with an estimated
transferability $\hat\mathcal{T}(\mathbf{des(c)}) less than 0.10 m$. This
represents a distance between the endpoint reached in simulation and
the one reached in reality lower than 10 cm. 


\subsubsection{Results}

\begin{figure*}
\centering
\includegraphics[width=\linewidth]{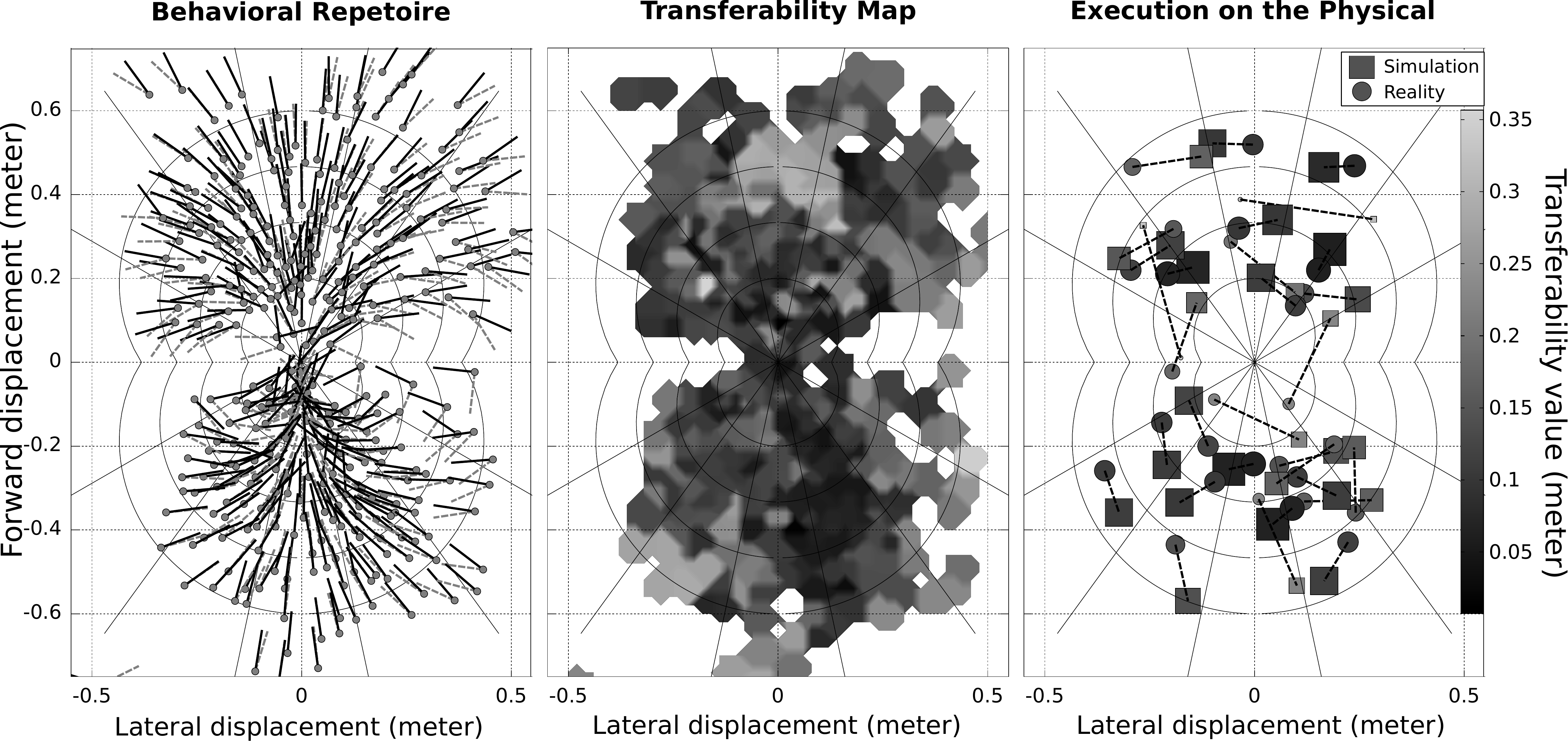}
\caption{Typical repertoire of controllers obtained with the \algoname{} algorithm.
(Left) The dots represent the endpoints of each controller. The solid lines are the final 
orientations of the robot while the dashed ones are the desired orientations. The angle between 
these two lines is the orientation error.
(Center) Transferability map. For each point of the reachable space, the estimated transferability of the nearest controller, within a radius of 5cm, is pictured.
(Right) Execution on the physical robot. The 30 selected controllers are pictured with square and their actual
endpoint with circles. The size and the color of the markers are proportional to their accuracy.
To select the tested controllers, the reachable space is split into 30 regions.
Their boundaries are defined by two lines at 60 degrees on each side of the robot and by
two curved frontiers that regroup all reachable points with a curvi-linear abscissa between 0.2 and 0.6 m.
These regions are then segmented into 15 parts for both the front and the rear of the robot.
All of these values are set from experimental observations of commonly reachable points.}
\label{fig:archive}
\end{figure*}

After 3000 iterations and 60 transfers, \algoname{} generates a
repertoire with a median number of $375$ controllers (min = $352$, max
= $394$). This is achieved in approximately 2.5 hours. One of these
repertoires is pictured in figure~\ref{fig:archive}, left.  The
distribution of the controllers' endpoints follows the same pattern as
in the virtual experiments: they cover both the front and the rear of the
robot, but not the lateral sides.  Here again, these limits are not
explicitly defined, they are autonomously discovered
by the algorithm.

Similarly to the experiments on the virtual robot, the majority of the controllers have a good final orientation and only the peripheries of the 
repertoire have a distinguishable orientation error.
\algoname{} successfully pushes the repertoire of controllers towards controllers with a good 
quality score and thus following the desired trajectories.

From these results we can draw the same conclusion as with the
previous experiment: the difficulty of accessing peripheral regions
explains the comparatively poor performances of controllers from these
parts of archive. The large distance to the starting point or the
complexity of the required trajectory meets the limits of the employed
controllers.  

The transferability map (Fig. \ref{fig:archive}, center) shows that the majority of the controllers 
have an estimated value lower than 15cm (dark regions). Nevertheless, some regions are deemed non-transferable (light regions). 
These regions are situated in the peripheries too, but are also in circumscribed areas inside of the reachable space.
Their occurrence in the peripheries have the same reasons as for the orientation (section \ref{sec:virtual_results}), but those inside the
reachable space show that the algorithm failed to find transferable controllers in few specific regions.
This happens when the performed transfers  do not allow the algorithm to infer transferable controllers. To overcome this issue, 
different selection heuristics and transfer frequencies will be considered in future work. 

In order to evaluate the hundreds of behaviors contained in the repertoires on the physical robot, we select 30 controllers in 
each repertoire of the $5$ runs.
The selection is made by splitting the space into 30 areas (Fig.~\ref{fig:archive}) and 
selecting the controllers with the best estimated transferability in each area.

\begin{figure}
\centering
  \includegraphics[width=0.75\columnwidth]{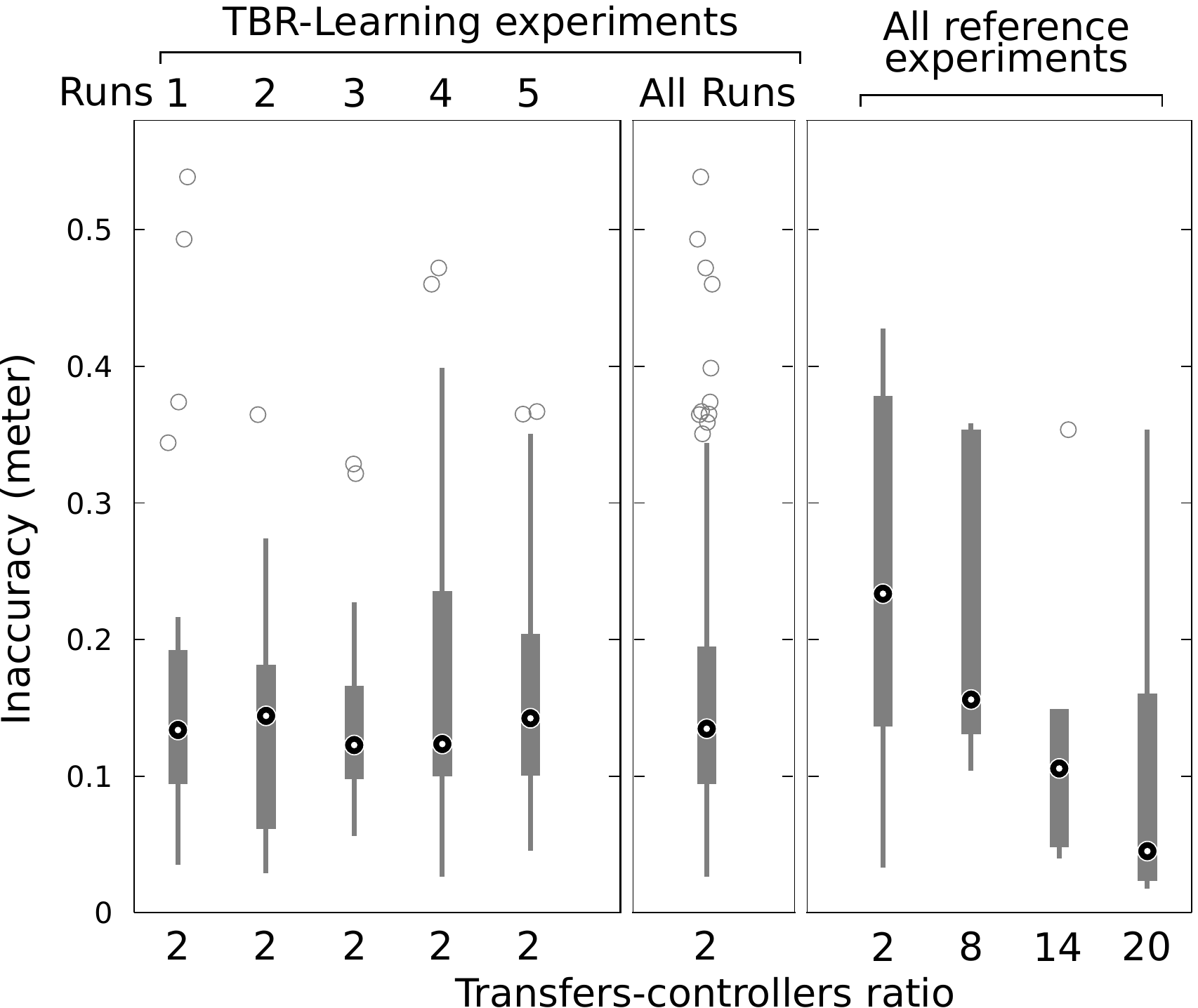}
\caption{Accuracy of the controllers. The accuracy is measured as the distance between the
endpoint reached by the \emph{physical robot} and the one reached in simulation (30 points for each run, see text). The results of the 
\algoname{} experiments are, for each run, separately pictured (Left)
and also combined for an overall point of view (Center). The performances of the reference experiments
are plotted according to the number of transfers performed (Right). In both cases, one transfer is performed
every 50 iterations of the algorithm.}
\label{fig:boxplot}
\end{figure}

Most of these controllers have an actual transferability value lower than $15$ cm (Fig.~\ref{fig:boxplot}, left), which 
is consistent with the observations of the transferability map (Fig.~\ref{fig:archive}, center) and not very large once taken 
into consideration the SLAM precision, the size of the robot and the looseness in the joints.
Over all the runs, the median accuracy of the controllers is $13.5$ cm (Fig.~\ref{fig:boxplot}, center). Nevertheless, every run
presents outliers, i.e. controllers with a very bad actual transferability value, which originate from
regions that the transferability function does not correctly approximate. 

In order to compare the efficiency of our approach to the reference
experiment, we use the ``transfers-controllers ratio'', that is the
number of performed transfers divided by the number of produced
controllers at the end of the evolutionary process.  For instance, if we reduce the
produced behavioral repertoires to the 30 tested controllers, this ratio is equal to $60 / 30 = 2$ for the \algoname{} experiments, since we performed 60 transfers.

The performances of the control experiments depend on the number of
performed transfers (Fig. \ref{fig:boxplot}, right) and thus on this
ratio. For an equal ratio, the reference experiments are $74\%$ less
accurate than \algoname{} ($13.5$ cm vs. $23.4$ cm, p-value=
$0.12$ with the Wilcoxon ranksum test), while the accuracies of
both experiments are not statistically different ($13.5$ cm
vs. $15.6$ cm and $10.6$ cm, p-value= $0.23$ and respectively $0.35$) if the
reference algorithm uses from 8 to 14 transfers to learn one
controller (i.e. a process 4 to 7 times longer).  The reference
experiment only takes advantage of its target specialisation when 20
transfers are performed.  With a transfers-controllers ratio equals to
20, the accuracy of the reference controllers outperforms the
controllers generated with the \algoname{} algorithm ($13.5$ cm vs
$4.5$ cm, p-value= $0.06$).  Nevertheless, with such a high ratio, the
reference experiment only generates $3$ controllers, while our
approach generates $30$ of them with the same running time (60
transfers and 3000 generations).

\begin{figure}
\centering
  \includegraphics[width=0.9\columnwidth]{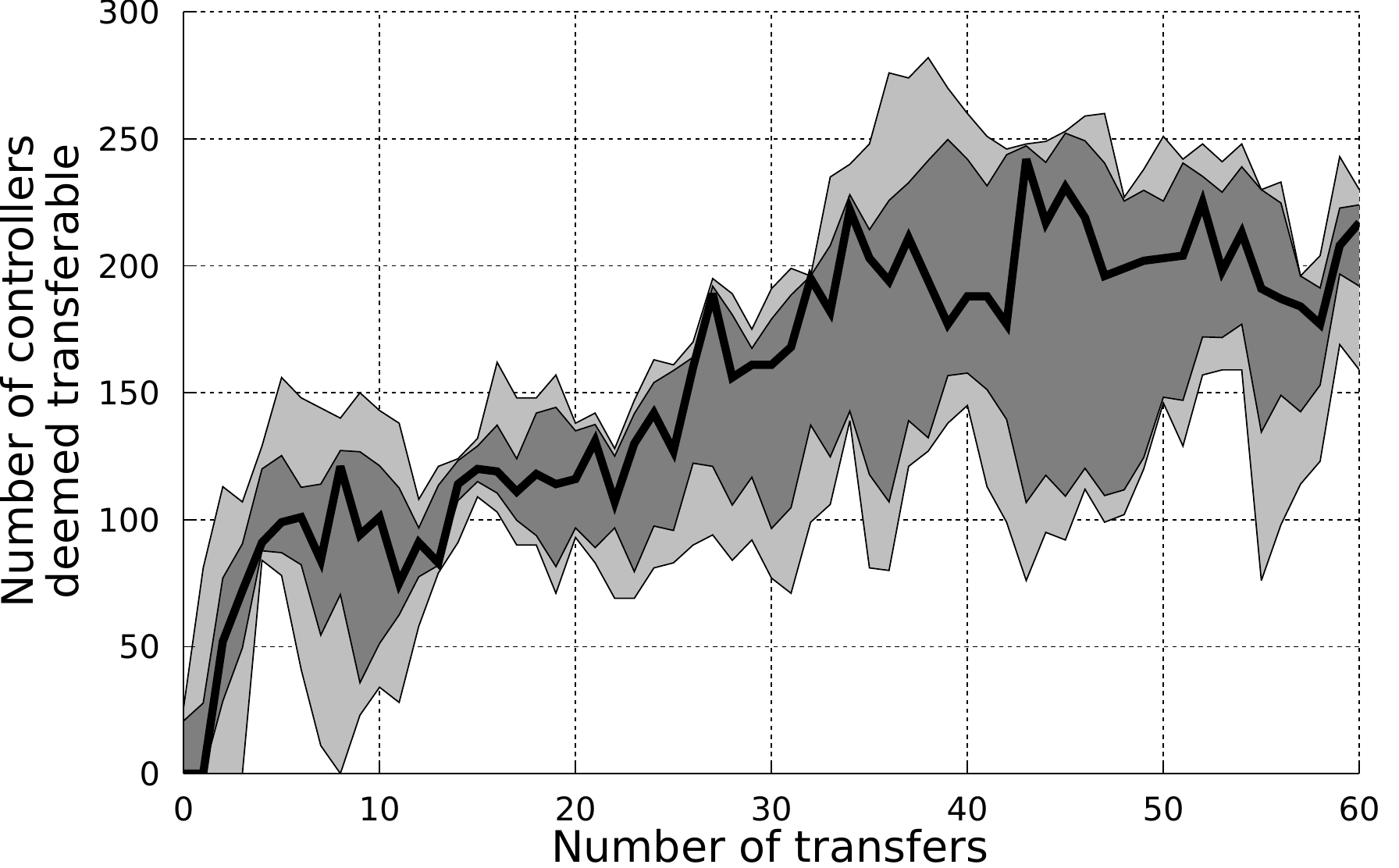}
\caption{Evolution of the number of controllers deemed transferable. At each transfer (i.e. every 50 iterations), the 
number of controllers with an estimated transferability lower than 15cm is pictured for all the 5 runs.
The bold black line represents the median value, the dark region the first and third quartile and the light
one the lower and upper bounds. The variability of the curve is due to the periodic transfers, which 
update the transferability function and thus the estimated transferability values.}
\label{fig:transferable}
\end{figure}

We previously only considered the 30 post evaluated controllers, whereas \algoname{} generates
several hundreds of them. 
After $60$ transfers, the repertoires contain a median number of 217 controllers that have an estimated transferability 
lower than $0.15$ m (Fig.~\ref{fig:transferable}). 
The previous results show that more than $50\%$ of the tested controllers have an actual transferability value lower than $0.15$ m and $75\%$ lower than $0.20$ m.
We can consequently extrapolate that between 100 and 150 controllers are exploitable in a typical behavioral repertoire. 
Once taking into consideration all these controllers, the transfers-controllers ratio of the \algoname{} experiments falls between $0.4$ and $0.6$ and 
thus our approach is about $25$ times faster than the reference experiment, for a similar accuracy.

\section{Conclusion and Discussion}

To our knowledge, \algoname{} is the first algorithm designed to
generate a large number of efficient gaits without requiring to
learn each of them separately, or to test complex controllers for each
direction. In addition, our experiments only rely on internal,
embedded measurements, which is critical for autonomy, but not
considered in most previous studies~(e.g., \cite{kohl2004policy,
  zykov2004evolving, chernova2004evolutionary, Yosinski2011,
  mahdavi2006innately}).

We evaluated our method on two experiments, one in simulation and one
 with  a physical hexapod robot.  With these experiments, we
showed that, thanks to its ability to recycle solutions usually wasted
by classic evolutionary algorithm, \algoname{} generate behavioral
repertoires faster than by evolving each solution separately. We also
showed that the archive management allows it to generate behavioral
repertoire with a significantly higher quality than the Novelty Search
algorithm \citep{lehman2011abandoning}.

With the \algoname{} algorithm, our physical hexapod robot was able to
learn several hundreds of controllers with only 60 transfers of 3
seconds on the robot, which was achieved in $2.5$ hours (including
computation time for evolution and the SLAM algorithm). The
repartition of these controllers over all the reachable space has been
autonomously inferred by the algorithm according to the abilities of
the robot.  Our experiments also showed that our method is about $25$
times faster than learning each controller separately.

\begin{figure}
\centering
\includegraphics[width=0.85\columnwidth]{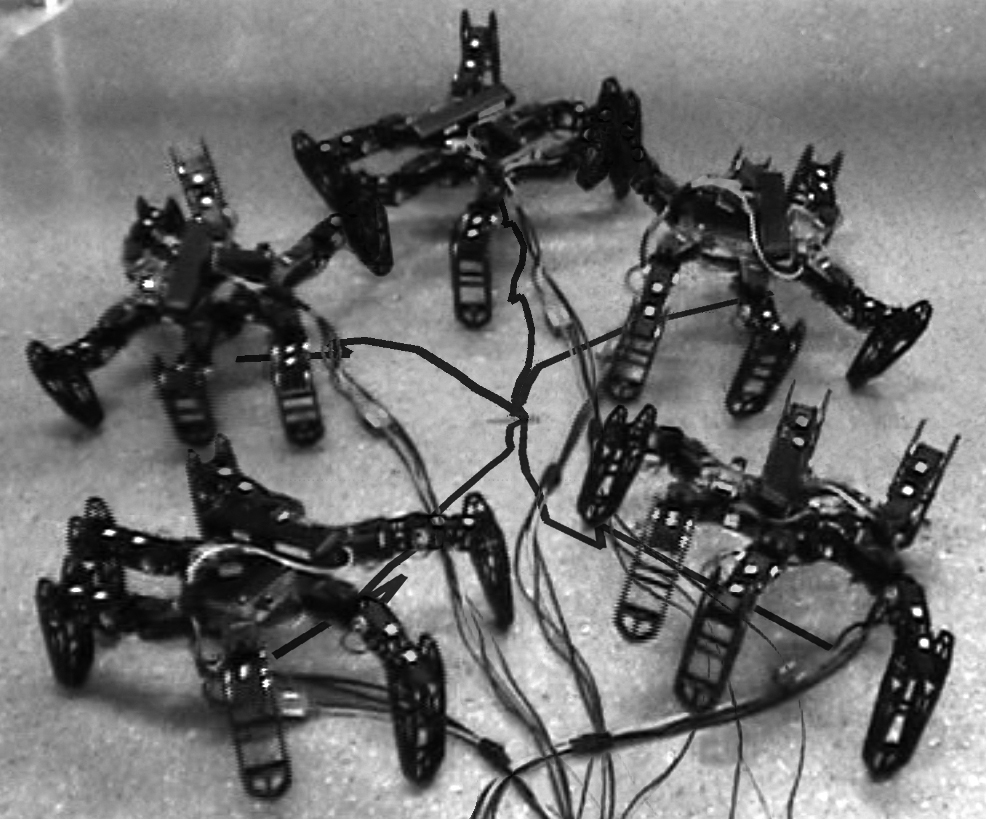}
\caption{Illustration of the results. 
These 5 typical trajectories  correspond to controllers obtained with \algoname{}
as recorded by the SLAM algorithm. The supplementary video shows a few other examples of controllers.}
\label{fig:mixture}
\end{figure}

Overall, these experiments show that \emph{the \algoname{} algorithm
  is a powerful method for learning multiple tasks with only several
  dozens of tests on the physical robot}. Figure~\ref{fig:mixture} and
the supplementary video illustrate the resulting ability of the robot
to walk in every direction. In the footsteps of Novelty Search, this
new algorithm thus highlights that evolutionary robotics can be more
than black-box optimization~\citep{Doncieux2014}: evolution can
simultaneously optimize in many niches, each of them corresponding to
a different, but high-performing, behavior.


In future work, we plan to investigate the generation of behavioral
repertoires in an environment with obstacles.  Selecting the transfers
according to the presence of obstacles might enable the robot to avoid
them during the learning process. This ability is rarely considered in
this kind of learning problem: most of the time, the robot is placed
in an empty space or manually replaced in a starting point (for
example \cite{kohl2004policy,zykov2004evolving,berenson2005hardware},
and the present work). This would be a step forward in autonomous
learning in robotics. 


Both the Transferability approach and the Novelty Search with Local
Competition do not put any assumption on the type of controller or
genotype employed. For example, the Transferability approach has been
used to evolve the parameters of Central Pattern
Generators~\citep{oliveiraoptimization} or those of an Artificial
Neural Networks ~\citep{koos2011transferability}, and the Novelty
Search algorithm has been employed on plastic artificial neural
encoded with NEAT~\citep{risi2010evolving,lehman2011abandoning} and on
graph-based virtual creatures~\citep{lehman2011evolving}. Similarly,
since the \algoname{} is the combination of these two algorithms, it
can also be used with any type of genotype or controller. In future
work, we will therefore investigate more sophisticated genotypes and
phenotypes like, for instance, neural networks encoded with
HyperNEAT~\citep{stanley2009hypercube,Clune2011,taraporecomparing,tarapore2014evolvability},
or oscillators encoded by compositional pattern-producing networks
(CPPN), like
SUPGs~\citep{morse2013single,taraporecomparing,tarapore2014evolvability}. Nevertheless,
such advanced controllers can use feedback to change their behavior
according to their sensors, and understanding how feedback-driven
controllers and a repertoire-based approach can be elegantly combined
is an open question.

The \algoname{} puts also no assumption on the type of robot and
  it would be interesting to see the abilities of the algorithm on
  more challenging robots like quadrupedal or bipedal robots, where the
  stability is more critical than with the hexapod robot.

The ability of \algoname{} to autonomously infer the possible actions
of the robot makes this algorithm a relevant tool for developmental
robotics \citep{lungarella2003developmental}. With our methods, the
robot progressively discovers its abilities and then perfects
them. This process is similar to the ``artificial curiosity''
algorithms in developmental robotics~\citep{Barto2004,Oudeyer2004},
which make robots autonomously discover their abilities by exploring
their behavioral space. It will be relevant to study the links between
these approaches and our algorithm, which come from different branches
of artificial intelligence. For example, can we build a behavioral
repertoire thanks to the artificial curiosity? Or, can we see the
novelty search aspects of \algoname{} like a curiosity process? Which
of these two approaches is less affected by the curse of
dimensionality?

\section{Acknowledgments}
This work has been funded by the ANR Creadapt project {\small (ANR-12-JS03-0009)} and a DGA/UPMC scholarship for A.C.

\section*{APPRENDIX}
\label{section:parameters}
The source-code of our experiments and a supplementary video
can be downloaded from: \url{http://pages.isir.upmc.fr/evorob_db}

\begin{itemize}
\item Parameters used for the experiments on the virtual robot:
\begin{itemize}
\item \algoname{}, Novelty Search and NS with Local Competition experiments:
\begin{itemize}
\item Population size: 100 individuals
\item Number of generations: 10 000
\item Mutation rate: 10\% on each parameters
\item Crossover: disabled
\item $\rho$: 0.10 m
\item $\rho$ variation: none
\item $k$:15
\end{itemize}

\item ``Nearest'' and ``Orientation'' control experiments:
\begin{itemize}
\item Population size : 100 individuals
\item Number of generations : 50 0000 (100 * 500)
\item Mutation rate : 10\% on each parameters
\item Crossover : disabled
\end{itemize}
\end{itemize}

\item Parameters used for the experiments on the physical robot:
\begin{itemize}
\item \algoname{} and the control experiment:
\begin{itemize}
\item Population size: 100 individuals 
\item Number of generations: 3 000 generations 
\item Mutation: 10\% on each parameters 
\item Crossover: disabled
\item $\rho$: 0.10 m 
\item $\rho$ variation: none
\item Transfer period: 50 iterations
\item $\tau$: -0.05 m
\end{itemize}
\end{itemize}
\end{itemize}

\bibliographystyle{apalike}
\bibliography{tbr_evolution.bib}

\end{document}